# Deep reflective reasoning in interdependence constrained structured data extraction from clinical notes for digital health

Jingwei Huang[1], Kuroush Nezafati[1], Zhikai Chi[2], Ruichen Rong[1], Colin Treager[1], Tingyi Wanyan[1], Yueshuang Xu[1], Xiaowei Zhan[1], Patrick Leavey[3], Guanghua Xiao[1], Wenqi Shi[1*], Yang Xie[1*]

1. Quantitative Biomedical Research Center, Department of Health Data Science and Biostatistics, Peter O'Donnell School of Public Health, University of Texas Southwestern Medical Center, 5323 Harry Hines Blvd, Dallas, TX, USA 75390
2. Department of Pathology, University of Texas Southwestern Medical Center, 5323 Harry Hines Blvd, Dallas, TX, USA 75390
3 Department of Pediatrics, University of Texas Southwestern Medical Center, 5323 Harry Hines Blvd, Dallas, TX, USA 75390

* co-corresponding authors

Acknowledgements: This work was partially supported by National Institutes of Health under award numbers P50CA70907, R35GM136375, R01GM140012, R01GM141519, and U01AI169298; NCTN Operations Center Grant (U10CA180886), NCTN Statistics & Data Center Grant (U10CA180899), COG Biospecimen Bank Grant (U24CA196173), St. Baldrick's Foundation, Wipe Out Kids' Cancer foundation, Quad W Foundation, and the Cancer Prevention and Research Institute of Texas under award numbers RP230330 and RP240521.

## Abstract

Extracting structured information from clinical notes requires navigating a dense web of interdependent variables where the value of one variable logically constrains others. Existing Large Language Model (LLM)-based extraction pipelines often struggle to capture these dependencies, leading to clinically inconsistent outputs. We propose deep reflective reasoning, a large language model agent framework that iteratively self-critiques and revises structured outputs by checking consistency among variables, the input text, and retrieved domain knowledge, stopping when outputs converge. We extensively evaluate the proposed method in three diverse oncology applications: (1) On colorectal cancer synoptic reporting from gross descriptions (n=217), reflective reasoning improved average F1 across eight categorical synoptic variables from 0.828 to 0.911 and increased mean correct rate across four numeric variables from 0.806 to 0.895; (2) On Ewing sarcoma CD99 immunostaining pattern identification (n=200), the accuracy improved from 0.870 to 0.927; (3) On lung cancer tumor staging (n=100), tumor stage accuracy improved from 0.680 to 0.833 (pT: 0.842→0.884; pN: 0.885→0.948). The results demonstrate that deep reflective reasoning can systematically improve the reliability of LLM-based structured data extraction under interdependence constraints, enabling more consistent machine-operable clinical datasets and facilitating knowledge discovery with machine learning and data science towards digital health.

## Introduction

Structured data extraction from unstructured free text to enable machine-operable data and facilitate knowledge discovery with machine learning and data science is crucial for digital health[1,2]. Clinical notes contain much of the information needed for clinical decision-making, quality reporting, registry abstraction, and research, yet these data remain difficult to use at scale because they are embedded in free text. The most practical representation of extracted structure is often a set of key–value pairs, where each key corresponds to a variable of interest (e.g., tumor site, margin status, histologic type) and each value corresponds to a standardized category or numeric measurement, or a list of them. In healthcare, as in other fields, variables of interest are typically interdependent rather than independent. The rise of Large Language Models (LLMs) has revolutionized the AI field and is shifting the paradigms of biomedical research and healthcare provision[3-10]. LLMs have become a powerful tool for structured data extraction or information extraction (IE)[11-14]. However, it has been challenging to make LLMs to stay with the constraints of variable interdependence relationships among variables. For example, LLM extracted attributes {"Margin Status for Invasive Carcinoma": "Invasive carcinoma present at margin"}, {"Closest Margin to Invasive Carcinoma": "Radial (circumferential) or mesenteric"}, and {"Distance of Tumor from Closest Margin (cm)": 0.3} are inconsistent, because they cannot be true at the same time. In term of mathematic logic, the conjunction of these three propositions is false or leads to a contradiction.

In this study, we propose deep reflective reasoning, an LLM-agent framework designed to improve structured biomedical information extraction under interdependence constraints. The approach performs iterative self-critique and self-revision: the agent reviews its own predicted variable values, checks them for consistency among variables, the source text, and domain knowledge, and updates outputs until predictions converge. **Fig. 1&2** summarize the agent architecture and the end-to-end workflow for integrating this framework into synoptic reporting. Our main motivating application is to generate pathology synoptic reports from gross description of unstructured clinical notes in the field of colorectal cancer. For this application, our experiments achieved average accuracy of 0.9167, where accuracies of 4 important variables including Rectal Tumor Location, Macroscopic Tumor Perforation, and Margin Status are over 0.95. This work demonstrates the feasibility of using deep reflective reasoning to generate pre-synoptic data reports from free text gross description. To further demonstrate the effectiveness and generality of our approach, we also applied the method to two other medical applications: (a) identifying CD99 immunostaining patterns from Ewing Sarcoma free text pathology reports, scanned from paper documents; (b) estimating lung cancer tumor stage from free text clinical notes.

Colorectal cancer (CRC) poses a major global health challenge. CRC is the third most common cancer with about 1.9 million new cases annually and the second leading cause of cancer-related death with 0.9 million annually worldwide[15,16]. A concerning trend is the increasing incidence of CRC among younger adults under the age of 50[17,18], which signals the needs of strengthening CRC knowledge discovery on diagnosis, prognosis, effective treatment, and prevention. Synoptic reporting refers to structured surgical pathology reporting, composed of a list of required data elements (or variables) and predefined standardized values, to ensure data completeness, accuracy, consistency, scalable data capture, interoperability, and exchange, thus facilitating biomedical research and clinical service delivery[19,20]. Compared with traditional narrative format of pathology reports, synoptic reports can also better facilitate machine learning for medical knowledge discovery and broader AI applications such as intelligent clinical decision supporting. The problem of generating synoptic data values from narrative or free text of gross description is a challenging task in three unique ways: (1) the task requires proper understanding of complex anatomic structure of colon and rectum and their associated complex terms and alternative synonyms; (2) the text description given in clinical notes such as gross description is often non-standardized, inexplicit, vague, ambiguous, and incomplete; (3) the task needs reasoning capability to derive a synoptic value from complex text description of related tumor properties. This article presents our approach to tackling the challenges by using deep reflective reasoning with Large Language Model (LLM) agents. From the perspective of leveraging AI in medical practice, this research intends to reduce the burden on pathologists from tedious and laborious work of reading messy gross reports or other clinical notes and to focus on critical work, as illustrated in Fig. 2. We use a dataset of 217 CRC cases collected at the University of Texas Southwestern Medical Center (UTSW) between 2010 and 2020 for this research. Together, these studies demonstrate that deep reflective reasoning can systematically improve the reliability of LLM-based structured extraction in settings where multiple interdependent clinical variables must be inferred and reconciled, enabling more consistent machine-operable datasets for digital health applications.

# Results

In the following, we present the proposed reflective reasoning algorithm, overall performance of colorectal cancer structured pre-synoptic data generation from gross description, reflection performance analysis, and applications to Ewing sarcoma CD99 pattern identification and lung cancer TNM staging.

## Deep reflective reasoning algorithm

Deep reflective reasoning is a multi-round iterative process that facilitates adjusting variable-value assignments for consistency through repeated reflections, conducting self-review and self-correction to ensure consistency among output variables, input text, and knowledge embeddings.

Reflective reasoning starts from a baseline prediction to generate a sequence of predictions, updating values of some variables for each case in the input dataset, till a case is converged or reaches a maximum round of reflections. As illustrated in **Fig. 1b**, in each round, it reviews the variable values bound in the last *p* rounds of reflections to inform the current reflection. For each case, if the reflection has not converged, a prompt is constructed using the prompt template, the input text of current case, and the reviewed attributes from previous rounds. This prompt is then used to query a LLM agent for new variable values. The LLM agent engages in self-criticism, cross-checking the consistency of its output variables with the input text and knowledge embeddings, thereby making self-corrections as necessary. If these new values match those from the previous round, the case is considered converged; otherwise, it continues to the next round of reflection. The process stops either when all cases have converged or when the maximum number of rounds *m* is reached. The output is a comprehensive collection of attributes for all cases across all rounds of reflection, providing a rich dataset for further analysis. We present the algorithm of deep reflective reasoning in supplementary **Fig S6**.

## Overall performance

We provide the performance of the final reflective reasoning over one-by-one query with RAG as follows. **Table 1 (a)** shows the overall performance on 8 categorical variables and the average performance metrics. The comprehensive score for all variables, average F1 score, is 0.9112. The number following each variable is the number of classes or categorical values that variable has. We provide the whole list of classes for each variable as a JSON file in the supplementary materials. “Margin Status for Invasive Carcinoma” (with 5 classes) achieved best performance with accuracy of 0.9697 and F1 score of 0.9707. “Tumor Extent” (with 9 classes) achieved an accuracy of 0.8198 and F1 score of 0.7933, which is the lowest performance score among all variables. **Table 1 (b)** presents the overall performance on 4 numeric variables. Correct rate is defined as the number of correctly predicted cases divided by the total number of cases evaluated for the variable. Correct rate is equivalent to accuracy in the sense of treating each specific numeric value as a class. A numeric prediction is regarded as “correct” if the predicted number exactly equals to the ground truth value. In the evaluation, we excluded cases where the ground truth of synoptic data conflicts with the real gross description text, or the ground truth is not precisely defined, such as values like “Cannot be determined”, “Other”, “Not applicable” and so on. The rate of cases evaluated is given as “coverage”. The coverage for numeric variables is lower, because in many pathology reports, the gross description did not provide specific values for some numeric synoptic data elements.

“Tumor Site” is a critical variable in the synoptic data generation process. In additional to the challenges addressed in the introduction section, another aspect of complexity for tumor site is that it has 12 elementary classes, and a tumor could appear in the intersection of two adjacent colorectal sections. “Tumor Site” achieved accuracy of 0.8582 and F1 score of 0.8484. **Fig. 3a** shows the confusion matrix of tumor site. From our experiments, we observed that LLMs frequently struggled with accuracy around ascending colon, cecum and ileocecal valve and around rectum, rectosigmoid, and sigmoid colon. In Fig. 3a, we merged rectosigmoid and rectum together as a single class, because rectosigmoid is virtually just a concept to indicating the boundary of rectum and sigmoid, and tumor site is assigned to rectosigmoid if and only if differentiation between rectum and sigmoid according is not possible[21]. In the confusion matrix, the most significant false prediction is around the subtle area of cecum and ascending colon, where LLM mistakenly classified 3 ascending colon cases as cecum. Overall, the number of false predictions is generally small. Later in Fig. 5, we will use a case as example to reveal the complexity of the problem and the reasoning.

**Fig. 3b ~ 3g** present the confusion matrixes of “Rectal Tumor Location”, “Macroscopic Tumor Perforation”, “Tumor Extent”, “Macroscopic Tumor Perforation”, “Margin Status for Invasive Carcinoma”, and “Closest Margin to Invasive Carcinoma”. Among them, “Margin Status for Invasive Carcinoma” (with 5 classes) achieved best performance with accuracy of 0.9697 and F1 score of 0.9707. “Tumor Extent” (with 9 variables) got accuracy of 0.8198 and F1 score of 0.7933, which is the lowest performance score among all categorical variables. This score reflects that the task for tumor extent is more challenging. As we will see later in case analysis, LLM struggled with complex anatomic structure and associated terminology of colon and rectum as well as their adjacent structures. As a result, in some scenarios, LLM failed to decern the semantics of some standard values for tumor extent, particularly, "Invades through muscularis propria into the pericolonic or perirectal tissue", "Invades visceral peritoneum (including tumor continuous with serosal surface through area of inflammation)", and "Directly invades or adheres to adjacent structure(s)". The case analysis illustrated in Fig. 4c provides more detailed analysis.

**Fig. 3h ~ 3k** present the estimated (y-axis) vs the ground truth (x-axis) of tumor size, distance of tumor to the closest margin, distance to radial margin, and distance to distal margin.

## Reflection performance analysis

Using the above algorithm and the result of one-by-one query with RAG as the baseline, we conducted 16 rounds of reflections. **Fig. 4a** presents the average F1 score for categorical synoptic variables, the average correct rate for numerical variables, and the number of cases revised in each round of reflection. As the figure shows, after round 3, the reflective reasoning converged to average correct rate of 0.8953 for all numeric variables; after round 10, the reflective reasoning converged to two average F1 scores, 0.9112, and 0.9061, for all categorical variables; after round 10, the reflection converged with 193 cases but kept revising the rest 24 cases. **Fig. 4b** illustrates the performance improvement and convergence with the rounds of reflections. After 10 rounds of reflection, the reflective reasoning converged for all numeric variable and most categorical variables but oscillated only on one or two standard values of 24 categorical variables. **Fig. 4c** provides an in-depth investigation about reflection process by showcasing a specific case (which is one of 24 cases not completely converged), where the reflection oscillates between two standard values of “Tumor Extent”, and all other synoptic variables are converged in the reflective reasoning. In this case, the reflective reasoning revised the standard value of synoptic variable “Margin Status for Invasive Carcinoma” from “Invasive carcinoma present at margin” to “All margins negative for invasive carcinoma” in round 10 and converged to this revised value in the followed rounds. Starting from round 10, the reflective reasoning begins to oscillate between two standard values of “Directly invades or adheres to adjacent structure(s)” and “Invades through muscularis propria into the pericolonic or perirectal tissue” for synoptic variable “Tumor Extent”, reflecting LLM having high uncertainty. As Fig 4a shows, there are 24 cases not completely converged and oscillating with one or two variables on two standard values, just like the example shows.

From our observations in the experiments, deep reflective reasoning may converge to two type of states: (1) a unique state, where each variable converges to a unique value; (2) an attractor, or a ring of several states, where at least one variable keeps switching from one value to another within two (or more) values, that is, LLM output keeps “walking” in a circle. To ensure having the same LLM output for the same input, we have set LLM’s parameter temperature as 0.

Therefore, when deep reflective reasoning converges to either a unique state or a more dynamic attractor, LLM out will stay with those states rather than generate new states.

## Comparison of alternative LLM solutions

**Table 2** compares the performance of different alternative LLM solutions, including,

1) one-by-one query without knowledge embedding
2) one-by-one query with RAG
3) reflection on one-by-one query with RAG (round 10, best in the converging states)
4) reflection on one-by-one query with RAG (round 11, worst in the converging states)

From the table, we can see that the average F1 score of one-by-one query with RAG is 0.07 higher than the one without RAG, reflecting the effectiveness of RAG. The reflective reasoning further improves the average F1 score by 0.0835, or minimally 0.0784, comparing with one-by-one query with RAG, reflecting the effectiveness of the reflection.

## Case analysis

**Fig. 5** presents our error analysis on reasoning about tumor site, a critical synoptic data item. We conduct the analysis with two cases. **Fig. 5a** illustrate an interesting scenario with case R0014. In the illustration of colon and rectum[22], the length of each colorectal section is from CAP guideline[21] (B. Anatomic Sites). The gross description about the tumor site is a bit complex and requires careful interpretation of anatomical landmarks. The LLM (Llama3.3-70b) focused on partial information and overlooked a critical component of the gross description: "8.5 cm from the proximal margin, 12.0 cm from the distal margin." This positional detail, when analyzed in full context, provides a clearer indication of the tumor's location along the colon. However, the phrase "located within the proximal aspect of the ascending colon" combined with the 8.5cm proximal margin may have biased the LLM to conclude incorrectly that the tumor resides within the ascending colon. This confusion highlights a core limitation: the model's insufficient understanding of gastrointestinal anatomy and surgical specimen orientation. Moreover, the LLM's reflections failed to correct this initial misinterpretation, likely due to the core limitation and lack of deep anatomical reasoning. This case underscores the importance of integrating domain-specific anatomical knowledge into knowledge engineering with LLMs. The task of accurately inferring tumor site from clinical notes is particularly challenging, as it demands not only linguistic comprehension but also a robust spatial understanding of organ structures and surgical practices. Without such understanding, even advanced language models may consistently misclassify or oversimplify anatomical descriptions.

We present another case, R044, in **Fig. 5b**. The gross description did not explicitly specify which colon section the tumor site is. The descriptions about specimen and margins imply the tumor located within ascending colon. According to CAP guideline (B. Anatomic Sites)[21], the total length of cecum and ascending colon is about 29cm maximally. The specimen is 27cm, where ascending colon is between 18-20cm, and cecum is between 7-9cm. Given that distal margin is 5.5cm and proximal margin is 15cm, no matter what length the ascending colon / cecum has, the tumor should be within ascending colon. The LLM picked the relevant text from the lengthy gross description; however, the LLM failed to infer the tumor site correctly, probably due to misleading by partial information of "Within the cecum, there is a 1.5 cm perforation …", as well as the weakness in geometric and numeric reasoning in the context of complex anatomic structure. The LLM reflections failed to correct the error for the same weakness.

Next in **Fig. 6**, we switch the analysis to the effects of reflection by using case R0227 as an example. **Fig. 6a** shows that the gross description of multiple lesions and ileum being the largest part of the specimen may lead the LLM biased towards collecting evidence to support cecum as major tumor site in both the initial synoptic data estimation and the reflections. Actually, from the right colon specimen of 17cm with lesion "extends to distal margin" implies the tumor resides at ascending colon rather than cecum. **Fig. 6b** shows that, in the initial estimation of radial margin, the LLM collected the relevance evidence but confused with the concepts about lesion dimensions and radial margin; however, in reflection process, the LLM correctly identified and fixed the error. This error made in the initial estimation reveals that the LLM is struggling with the complex terms in gross description to understand the concepts correctly. In **Fig. 6c**, for distal margin, the LLM accurately captured the right evidence and made correct judgment in both initial estimation and the reflections. In **Fig. 6d**, for closest margin, due to previous error of treating 0.5cm as radial margin, the LLM made the same mistake to assess closest margin as 0.5cm in initial estimation. What even worse is that the LLM did not take the distal margin of zero into account in estimating closest margin, reflecting the poor understanding of the key concept. Fortunately, in reflective reasoning, the LLM correctly fixed the error. Finally, in **Fig. 6e**, we present the estimation of margin status. In the initial estimation, the LLM made wrong assessment of "All margins negative for invasive carcinoma", although the LLM considered all relevant evidence, including "… extends to the distal margin", which is somewhat ambiguous about whether to really involve distal margin or not, and the LLM argued that "extends to but does not involve the distal margin". In the reflection, the LLM focused on the core evidence and correctly fixed the previous errors.

## Deep reflective reasoning about CD99 immunostaining pattern in Ewing sarcoma

We demonstrate the effectiveness of deep reflective reasoning with a different application – CD99 immunostaining pattern identification in Ewing sarcoma (EWS), which is a rare bone cancer in young people. CD99 (also known as MIC2) is an integral membrane glycoprotein strongly and is diffusely expressed in nearly all EWS tumors, thus being considered an important factor for the diagnosis of Ewing sarcoma[23,24]. CD99 immunostaining pattern is characterized in combinations of two aspects: (1) membranous vs cytoplasmic, and (2) focal, patchy vs diffuse. To demonstrate the generality of deep reflective reasoning with different LLMs, in this application, we use an advanced reasoning LLM – gpt-5.1. **Fig. S4a** presents the performance metrics with confusion matrix of CD99 pattern identification by gpt-5.1without reflection. **Fig. S4b** presents the performance metrics with confusion matrix of CD99 pattern identification by gpt-5.1with reflection using back-view of 2 rounds (i.e. for each case, looking back two rounds of LLM output). The reflective reasoning converged at round 3. The deep reflective reasoning improves the accuracy from 0.8698 (without reflection) to 0.9271.

## Deep reflective reasoning about lung cancer TNM staging

The third application we select to demonstrate deep reflective reasoning is lung cancer TNM staging with clinical notes. By AJCC7, Tumor Stage is dependent on three variables primary tumor (pT), Regional Lymph Nodes (pN,) and Distant Metastasis (pM) as presented in **Fig. S5**.

pT depends on factors of tumor size, tumor extent, tumor nodules, and the extent of tumor associated atelectasis or obstructive pneumonitis, with more sophisticated relations intwined with anatomical structures. For this more complex problem, we use back-view of 10 rounds in reflective reasoning. We execute the reasoning 15 rounds and there are only 2 cases remaining unconverged. **Fig. S3** left panel shows the results of TNM staging estimation by Llama3.3 without reflection. The right panel shows the results by 15 rounds of deep reflective reasoning with back-view of 10 rounds. The deep reflective reasoning improves the accuracy of pT from 0.8421 to 0.8842, the accuracy of pN from 0.8854 to 0.9479, and the accuracy of tumor stage from 0.6796 to 0.8333.

## Discussion

This article presented our approach to utilizing deep reflective reasoning of LLM for structured data extraction from unstructured free text of clinical notes in three biomedical applications, including (1) generating structured pre-synoptic data from gross description of colorectal cancers, (2) identifying immunohistochemical (IHC) staining patterns from the scanned archived pathology reports of Ewing sarcoma, curated by the Children's Oncology Group over 19 years, and (3) assessing lung cancer TNM staging from clinical notes. Across these settings, reflective self-critique systematically improved accuracy over a one-pass extraction baseline and reduced clinically implausible contradictions among extracted variables, supporting the use of LLM agents to create more reliable machine-operable datasets for digital health.

Reflection[8,25-27] is an important mechanism for improving LLMs' performance. The deep reflective reasoning method presented in this article is different from most existing reflection frameworks such as DeepSeek R1[8], Reflexion[25], and CRITIC[27], which combine with reinforcement learning paradigm and use external feedback from environment. Our deep reflective reasoning approach don't need external/additional supervision signal for revision, instead we use self-correction by self-critics on the consistency between the response of the LLM and the given facts and knowledge, thus being purely self-reflection, an internal thinking process of review and revise till converged; and we feed LLM with multiple previous round outputs as a mechanism to compensate LLMs weakness of without backtracking. Usually, existing frameworks let an LLM agent reflection for a fixed number of rounds. However, different rounds typically lead to different results, thus imposing some randomness to the result. We attempt to conduct much longer reflective reasoning with multiple back-views and stop the reasoning when the reflection is converged, or at least till a dominating number (such as over 90%) of cases in question converged.

Our method is inspired by the following observations. With a transformer-based architecture and self-attention mechanisms[3], an LLM generates output through an autoregressive inference process where the model iteratively predicts the most probable next token in a sequence based on the statistical patterns and semantic relationships learned during pre-training with vast amount of text data. Therefore, essentially an LLM makes reasoning in a sense of statistical reasoning, just like a seasoned master to answer a question based on his rich experience-grounded intuition. Logic-based reasoning systems use heuristic search, such as tree-based depth-first or breath-first search strategy combined with "backtracking" mechanism to search for consistent binding of values to the requested variables from a large number of possible combinations of variable values. LLMs do not have mathematically rigor mechanisms to confine their search space within the specified constraints or implicitly implied constraints of logical relationships among the attributes of interest, the given facts, and the knowledge to use. As a result, LLMs may generate output that is inconsistent between the values of variables or inconsistent with facts or knowledge. However, it is possible to let LLMs to review or criticize the outputs by examining the possible inconsistencies and revise the output, thus improving the final output. While self-correction is an interesting and challenging mechanism for LLM agents, our research with the presented three applications shows that it is feasible for LLM agents to make self-correction through self-reflection without external feedback from the environment. Even for the advanced reasoning model such as gpt-5.1, there is still room for improvement through reflection.

From the medical application perspective, the results of our experiments with colorectal cancer show that it is feasible to utilize deep reflective reasoning for automatic generation of structured synoptic data from free text of gross description of colorectal cancers. This AI powered automation in the direction of digital health can simplify the workflow of pathology reporting, significantly reduce pathologists' time on the tedious work of reading frequently messy gross description, improve accuracy in handling the text information, and focus on studying the pathology slides.

CD99 (MIC2) plays a critical role in Ewing sarcoma diagnosis[23,24]. CD99 immunostaining patterns were expressed in various free text expression patterns, in addition to data acquisition challenges since pathology reports were collected by COG from multiple different institutions over 19 years, with different styles, and collected as scanned PDF documents with various qualities. We use GPT-5.1, an advanced reasoning LLM, to cope with the problem. By applying deep reflective reasoning, we improved the accuracy of CD99 pattern identification from 0.8698 to 0.9271. The high accuracy of CD99 pattern identification will enable EWS researchers to explore its importance in diagnosis, treatment or outcome prediction.

Our experiments with lung cancer TNM staging further reveals that deep reflective reasoning can help to improve the quality of structured data extraction from unstructured free text, particularly, for the problems with multiple interdependent variables of interest. The problem we address here is beyond simple data extraction from text where there are direct and explicit statements, instead there is the need of reasoning with domain-specific intertwined biomedical relations among variables. Using standardized structured data in healthcare is a crucial move for the purpose of enabling machine-operable data and facilitating knowledge discovery with machine learning and data science towards digital health.

Further research can go in several directions. First of all, reasoning is a core essential capability that an AI agent as either co-scientist or copilot for clinicians should have. We will extend our research in the direction of enhancing LLM agents reasoning capability via more effective knowledge engineering. Secondly, we will extend our LLM facilitated workflow of synoptic reporting to include image feature recognition from digital slides, to build a copilot for pathologists. Finally, we will apply deep reflective reasoning in disease diagnosis and prognosis by using structured data extracted from multiple sources such as encounter notes, progress notes, pathology reports, CT notes, MRI notes, and others.

In summary, deep reflective reasoning provides an effective, model-agnostic strategy to improve LLM-based structured information extraction in settings with multiple interdependent clinical variables. By promoting internally consistent, guideline-aligned outputs, it supports scalable creation of machine-operable clinical datasets and strengthens the foundation for downstream analytics and digital health applications.

# Methods

## Data and data processing

For our first application, we collected 217 colorectal cancer pathology reports across 2010 to 2020 from the UTSW hospitals and clinics. For our project of generating synoptic data from clinical notes, we extracted out the gross description section of the pathology reports. The pathologist in our team selected 12 synoptic variables that are possibly read from gross description. For those gross description, pathology specialty trained medical professionals annotated the ground truth by using CAP (College of American Pathologists) Cancer Template for colon and rectum[19]. Then, we use the ground truth of those gross description to evaluate the performance of our method of synoptic data automatic generation by LLM.

For the second application on CD99 staining pattern identification, we randomly selected 200 cases from 932 pathology reports of Ewing sarcoma patients collected and curated by COG. The EWS expert in the team manually annotated the ground truth, which we used for evaluating the performance in this application.

For the third application on lung cancer TNM staging, we randomly selected 100 cases from 852 lung cancer pathology reports that we curated. The 852 cases were selected from TCGA Pan-Lung Cancer dataset (https://www.cbioportal.org/study/summary?id=nsclc_tcga_broad_2016) and excluded some cases with very poor quality of scanning. The certified physicians under supervise of pathologist in our team manually annotated the 852 pathology reports with ground truth using AJCC7[28].

## Knowledge engineering for LLMs

The problems we targeted are all domain-specific, thus needing domain knowledge packages to support. In our architecture, we use an LLM as a reasoning engine, which owns "general" knowledge through pre-training with vast amount of various data. Domain knowledge acquisition for LLMs is significantly simpler and straightforward, compared with tradition knowledge engineering that need formally represent all knowledge needed. We can simply gather the relevant domain knowledge in its nature form – text files in a designated directory. To make those text form of knowledge work best with LLMs, we reorganize them to make them easier for LLMs to catch and easier to cite. For example, we organize the AJCC7 TNM staging guide as a set of rules, and each rule is numbered. We organize those knowledge packages as markdown files.

To use those knowledge packages, we use LlamaIndex (https://www.llamaindex.ai/) as tool for RAG (Retrieval-Augmented Generation[29-33]). The granularity of the text augmentation with RAG could be in fine-grained level (such as, word as unit) or coarse-grained (such as, paragraph as unit). In order to avoid missing the conditions for some propositions and keep the completeness of a piece of knowledge to be used, we use section (of markdown file) as unit in RAG.

## Reflective reasoning

We have presented the reflective reasoning in Fig.1b and Fig. S6. The data structure we used in the reflective reasoning process is as follows. We use a sequence of predictions with critiques to previous variable-value bindings (or assignments). Let L be the number of variables of interest to characterize each case. For variable $v_i$, let $v_1_val^{(k)}$ be the value bound to the variable at round $k$, $v_1_c^{(k)}$ be the critique LLM agent made about the inconsistency identified, and $v_1_e^{(k)}$be the explanation about the critique $v_1_c^{(k)}$ and the updated variable-value binding $v_1_val^{(k)}$.

After round $k$ ($k=1, \ldots, m$), the sequence for case j will be as follows.

$$case_j^{(0)} = (v_1_val^{(0)}, v_1_c^{(0)}, v_1_e^{(0)}, \ldots, v_i_val^{(0)}, v_i_c^{(0)}, v_i_e^{(0)}, \ldots, v_L_val^{(0)}, v_L_c^{(0)}, v_L_e^{(0)})$$

$$\ldots,$$

$$case_j^{(k)} = (v_1_val^{(k)}, v_1_c^{(k)}, v_1_e^{(k)}, \ldots, v_i_val^{(k)}, v_i_c^{(k)}, v_i_e^{(k)}, \ldots, v_L_val^{(k)}, v_L_c^{(k)}, v_L_e^{(k)})$$

$$\ldots,$$

$$case_n^{(k)} = (v_1_val^{(k)}, v_1_c^{(k)}, v_1_e^{(k)}, \ldots, v_i_val^{(k)}, v_i_c^{(k)}, v_i_e^{(k)}, \ldots, v_L_val^{(k)}, v_L_c^{(k)}, v_L_e^{(k)})$$

$case_j^{(k)}$ is the result of reflection using p-back view $case_j^{(k-1)}$, …, $case_j^{(k-p)}$. In $case_j^{(0)}$, $v_i_c^{(0)}$ is blank, as reflection not started yet. For those cases converged in reflective reasoning, the length of the sequence of reflections is less than the maximum number of rounds $m$. To get the whole picture of the result at reflection round $k$, if $case_j^{(k)}$ does not exist, that is, the case converged earlier at a previous round $h < k$, the prediction for case j needs to be traced back from $case_j^{(h)}$.

We provided small examples to LLM for helping identifying inconsistencies. Examples of inconsistencies are shown as follows.

## Examples of Type 1 Inconsistencies

Example 1.1:

{

...

"Procedure": "Value *Left hemicolectomy* contradicts the fact ['The specimen received in formalin properly labeled with the patient's name designated 'terminal ileum, cecum, ascending, and appendix''], because Left hemicolectomy is a surgical resection in the left but Specimens is in the right."

"Procedure - Degree of Inconsistency": 1.0,

...

}

## Examples of Type 2 Inconsistencies

Example 2.4: *Margin Status for Invasive Carcinoma = Invasive carcinoma present at margin*, *Closest Margin to Invasive Carcinoma = Radial (circumferential) or mesenteric*, and *Distance of Tumor from Closest Margin (cm) = 0.3* are inconsistent, because they cannot be true at the same time.

## References

1 World Health, O. *Global strategy on digital health 2020-2027*. (World Health Organization, 2025).

2 Abernethy, A. *et al.* The promise of digital health: then, now, and the future. *NAM perspectives* **2022**, 10-31478 (2022).

3 Vaswani, A. *et al.* Attention is all you need. *Advances in neural information processing systems* **30** (2017).

4 Radford, A., Narasimhan, K., Salimans, T. & Sutskever, I. Improving language understanding by generative pre-training. *OpenAI - Research* (2018).

5 Kaplan, J. *et al.* Scaling laws for neural language models. *arXiv preprint arXiv:2001.08361* (2020).

6 Brown, T. *et al.* Language Models are Few-Shot Learners. *NeurIPS 2020* **33**, 1877-1901 (2020).

7 Touvron, H. *et al.* LLaMA: Open and efficient foundation language models. *arXiv preprint arXiv:2302.13971* (2023).

8 Guo, D. *et al.* DeepSeek-R1 incentivizes reasoning in LLMs through reinforcement learning. *Nature* **645**, 633-638 (2025). https://doi.org/10.1038/s41586-025-09422-z

9 McDuff, D. *et al.* Towards accurate differential diagnosis with large language models. *Nature* **642**, 451-457 (2025). https://doi.org/10.1038/s41586-025-08869-4

10 Lu, C. *et al.* The ai scientist: Towards fully automated open-ended scientific discovery. *arXiv preprint arXiv:2408.06292* (2024).

11 Huang, J. *et al.* A critical assessment of using ChatGPT for extracting structured data from clinical notes. *npj Digital Medicine* **7**, 106 (2024). https://doi.org/10.1038/s41746-024-01079-8

12 Hein, D. *et al.* Iterative refinement and goal articulation to optimize large language models for clinical information extraction. *npj Digital Medicine* **8**, 301 (2025). https://doi.org/10.1038/s41746-025-01686-z

13 Chen, D., Alnassar, S. A., Avison, K. E., Huang, R. S. & Raman, S. Large Language Model Applications for Health Information Extraction in Oncology: Scoping Review. *JMIR Cancer* **11**, e65984 (2025). https://doi.org/10.2196/65984

14 Rajaganapathy, S. *et al.* Synoptic reporting by summarizing cancer pathology reports using large language models. *npj Health Systems* **2**, 11 (2025). https://doi.org/10.1038/s44401-025-00013-8

15 Bray, F. *et al.* Global cancer statistics 2022: GLOBOCAN estimates of incidence and mortality worldwide for 36 cancers in 185 countries. *CA: A Cancer Journal for Clinicians* **74**, 229-263 (2024). https://doi.org/https://doi.org/10.3322/caac.21834

16 Ferlay, J. *et al. Global Cancer Observatory: Cancer Today (GLOBOCAN 2022)*, <https://gco.iarc.who.int/media/globocan/factsheets/populations/900-world-fact-sheet.pdf> (2024).

17 Sung, H. *et al.* Colorectal cancer incidence trends in younger versus older adults: an analysis of population-based cancer registry data. *The Lancet Oncology* **26**, 51-63 (2025). https://doi.org/10.1016/S1470-2045(24)00600-4

18 ACS. *Key Statistics for Colorectal Cancer, American Cancer Society*, <https://www.cancer.org/content/dam/CRC/PDF/Public/8604.00.pdf> (2025).

19 Burgart, L. J., Chopp, W. V. & Jain, D. Protocol for the Examination of Resection Specimens From Patients With Primary Carcinoma of the Colon and Rectum (v 4.2.0.2). (College of American Pathologists (CAP), 2022).
20 Zheng, L. *et al.* Synoptic reporting for cancer surgery: Current requirements and future state. (American College of Surgeons, 2021).
21 Burgart, L. J., Chopp, W. V. & Jain, D. Protocol for the Examination of Resection Specimens From Patients With Primary Carcinoma of the Colon and Rectum. (College of American Pathologists (CAP), 2022).
22 NIDDK-NIH. Vol. 378 KB | 1192 x 1236 (ed 1) Drawing of the large intestine. The appendix, cecum, ascending colon, transverse colon, descending colon, sigmoid colon, rectum, and anus are labeled. (National Institute of Diabetes and Digestive and Kidney Diseases, National Institutes of Health., https://www.niddk.nih.gov/news/media-library/18248, 2025).
23 O'Neill, A. F. *et al.* Targeted Imaging of Ewing Sarcoma in Preclinical Models Using a 64Cu-Labeled Anti-CD99 Antibody. *Clinical Cancer Research* **20**, 678-687 (2014). https://doi.org/10.1158/1078-0432.CCR-13-1660
24 Manara, M. C. *et al.* Engagement of CD99 Activates Distinct Programs in Ewing Sarcoma and Macrophages. *Cancer Immunol Res* **12**, 247-260 (2024). https://doi.org/10.1158/2326-6066.CIR-23-0440
25 Shinn, N., Cassano, F., Gopinath, A., Narasimhan, K. & Yao, S. Reflexion: Language agents with verbal reinforcement learning. *Advances in Neural Information Processing Systems* **36**, 8634-8652 (2023).
26 Madaan, A. *et al.* Self-refine: Iterative refinement with self-feedback. *Advances in Neural Information Processing Systems* **36**, 46534-46594 (2023).
27 Gou, Z. *et al.* Critic: Large language models can self-correct with tool-interactive critiquing. *ICLR 2024 (arXiv:2305.11738)* (2024).
28 Edge, S. B. & American Joint Committee on Cancer, A. C. S. *AJCC cancer staging manual, 7th Ed.*, (Springer, 2010).
29 Lewis, P. *et al.* Retrieval-augmented generation for knowledge-intensive nlp tasks. *Advances in neural information processing systems* **33**, 9459-9474 (2020).
30 Jiang, J. *et al.* 7064-7074.
31 Yu, T., Zhang, S. & Feng, Y. Auto-rag: Autonomous retrieval-augmented generation for large language models. *arXiv preprint arXiv:2411.19443* (2024).
32 Huang, Y. & Huang, J. A survey on retrieval-augmented text generation for large language models. *arXiv preprint arXiv:2404.10981* (2024).
33 Zhang, Q. *et al.* A survey of graph retrieval-augmented generation for customized large language models. *arXiv preprint arXiv:2501.13958* (2025).

# Figures & Tables

## Fig. 1: Overall framework of LLM agents with deep reflective reasoning

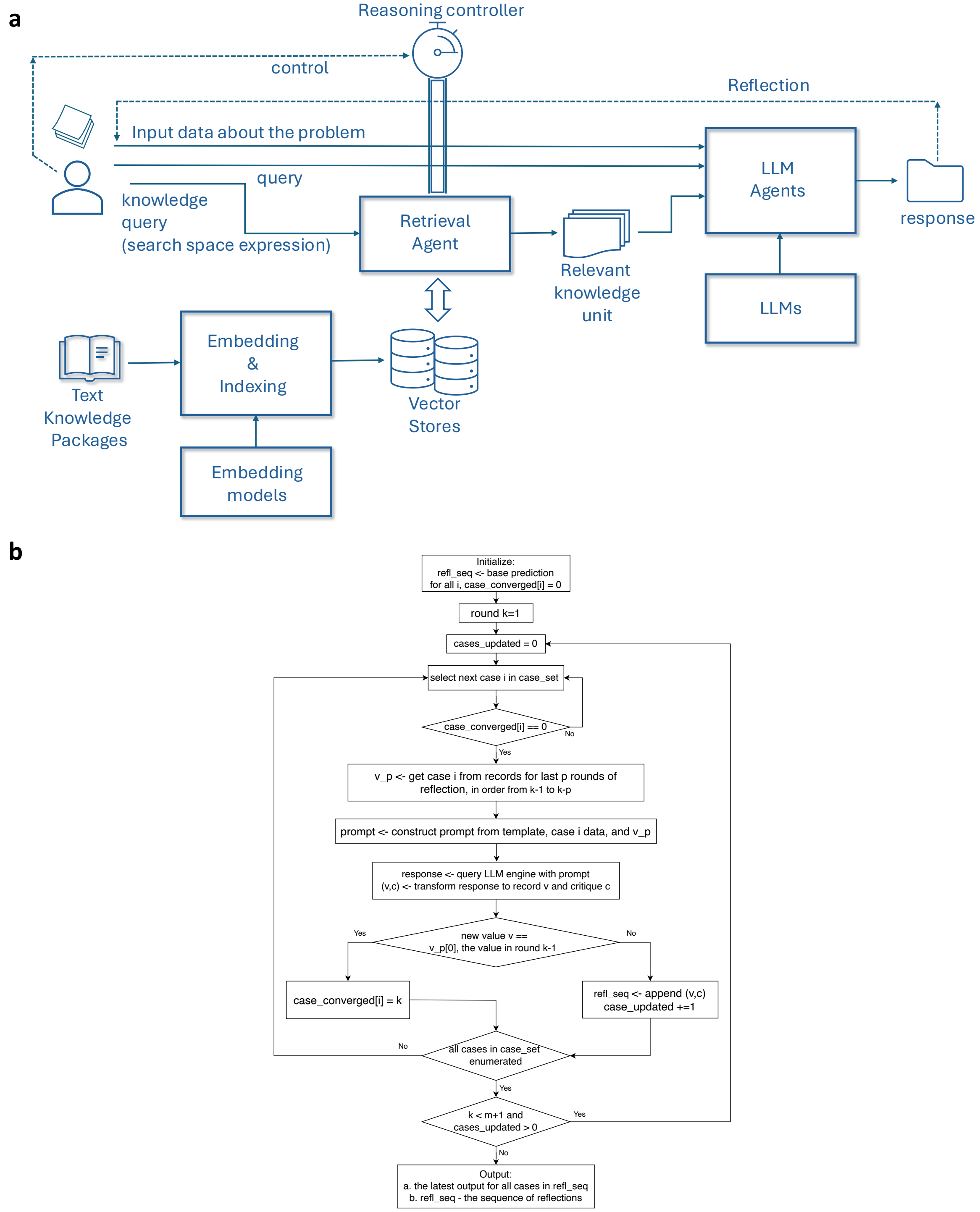

a. overflow of LLM agent with deep reflective reasoning; b. reflective reasoning algorithm

Fig. 2: Overall workflow of LLM agents supported synoptic reporting process

Overall workflow of LLM agents facilitated synoptic reporting process: In current medical practice, synoptic reports are created by pathologists, after they analyze and interpret pathological images and synthesize clinical notes. This research leverages LLM agent to extract synoptic attributes (if they are specified in clinical notes) or infer the attributes (if not specified), to create default values for the synoptic variables listed in CAP CRC protocol, thus reducing the burden on pathologists for reading the gross reports (or other clinical notes) in their workflow to make them focus on more important aspects.

Table 1: Performance metrics of reasoning about synoptic data

**a**

| **Synoptic variables**<br>(#): number of possible categories | **Accuracy** | **F1** | **Recall (weighted)** | **Precision (weighted)** | **Coverage** |
|---|---|---|---|---|---|
| Procedure (10) | 0.8957 | 0.9124 | 0.8957 | 0.9514 | 0.7512 |
| Macroscopic Evaluation of Mesorectum (6) | 0.9588 | 0.9573 | 0.9588 | 0.9609 | 0.7834 |
| Tumor Site (12) | 0.8582 | 0.8484 | 0.8582 | 0.8580 | 0.6498 |
| Rectal Tumor Location (4) | 0.9675 | 0.9586 | 0.9675 | 0.9607 | 0.7097 |
| Tumor Extent (9) | 0.8198 | 0.7933 | 0.8198 | 0.7705 | 0.5115 |
| Macroscopic Tumor Perforation (3) | 0.9610 | 0.9576 | 0.9610 | 0.9572 | 0.9447 |
| Closest Margin to Invasive Carcinoma (8) | 0.8986 | 0.8912 | 0.8986 | 0.8942 | 0.6359 |
| Margin Status for Invasive Carcinoma (5) | 0.9697 | 0.9707 | 0.9697 | 0.9722 | 0.6083 |
| **Average** | 0.9162 | 0.9112 | 0.9162 | 0.9156 | |

**b**

| **Synoptic variables** | **Correct Rate** | **MAE** | **RMSE** | **Coverage** |
|---|---|---|---|---|
| Tumor Size - Greatest dimension (cm) | 0.9680 | 0.1079 | 0.7094 | 0.8756 |
| Distance of Tumor from Radial Margin (cm) | 0.8670 | 0.5663 | 1.7153 | 0.4009 |
| Distance of Tumor from Distal Margin (cm) | 0.9760 | 0.1207 | 0.8367 | 0.3779 |
| Distance of Tumor from Closest Margin (cm) | 0.7700 | 0.8469 | 2.0639 | 0.5253 |
| Average | 0.8953 | 0.4104 | 1.3313 | |

a. Metrics for categorical synoptic variables; b. metrics for synoptic numeric variables.

Fig. 3: Confusion matrix for categorical variables and point plots for numeric variables

**a**

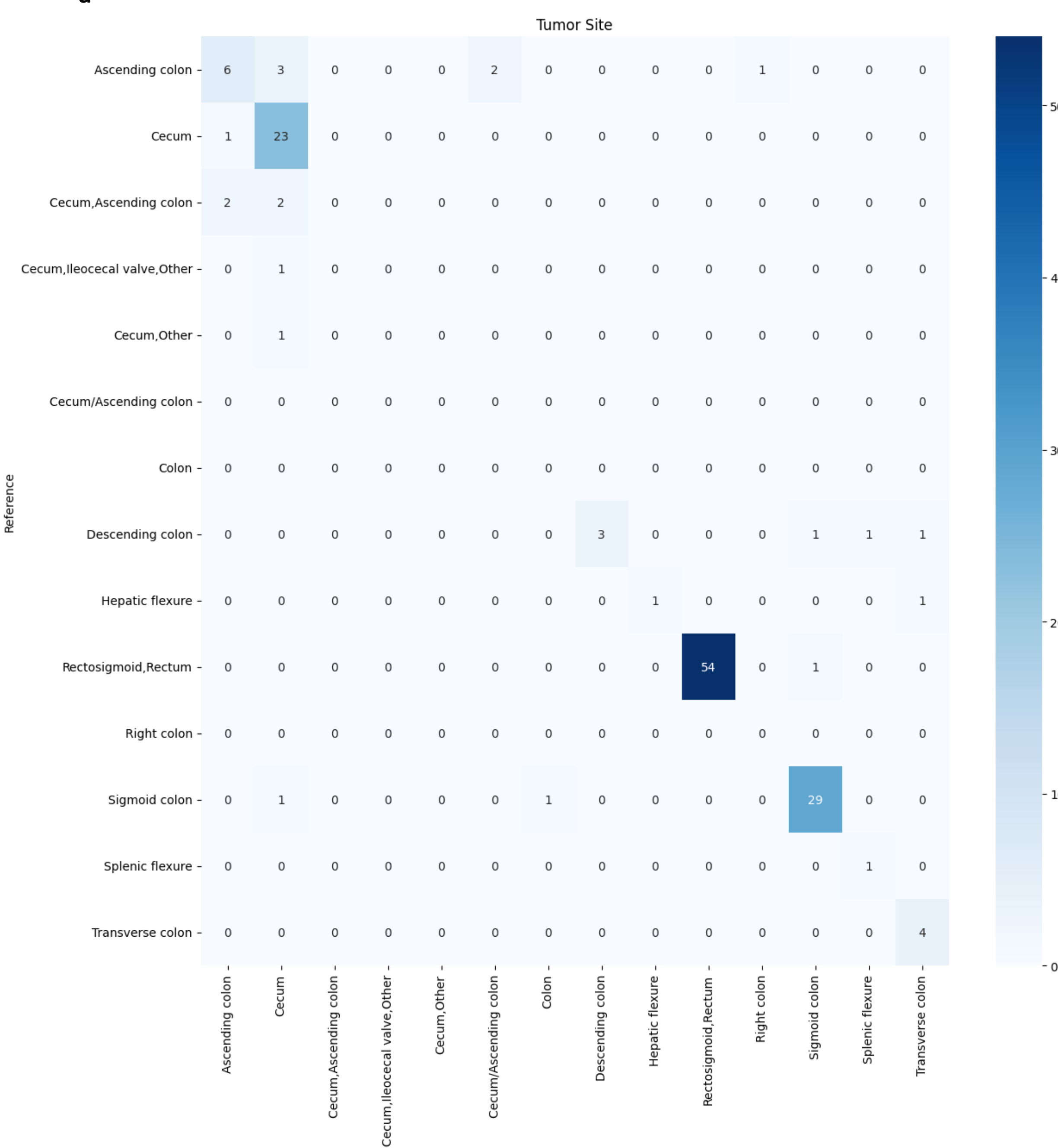

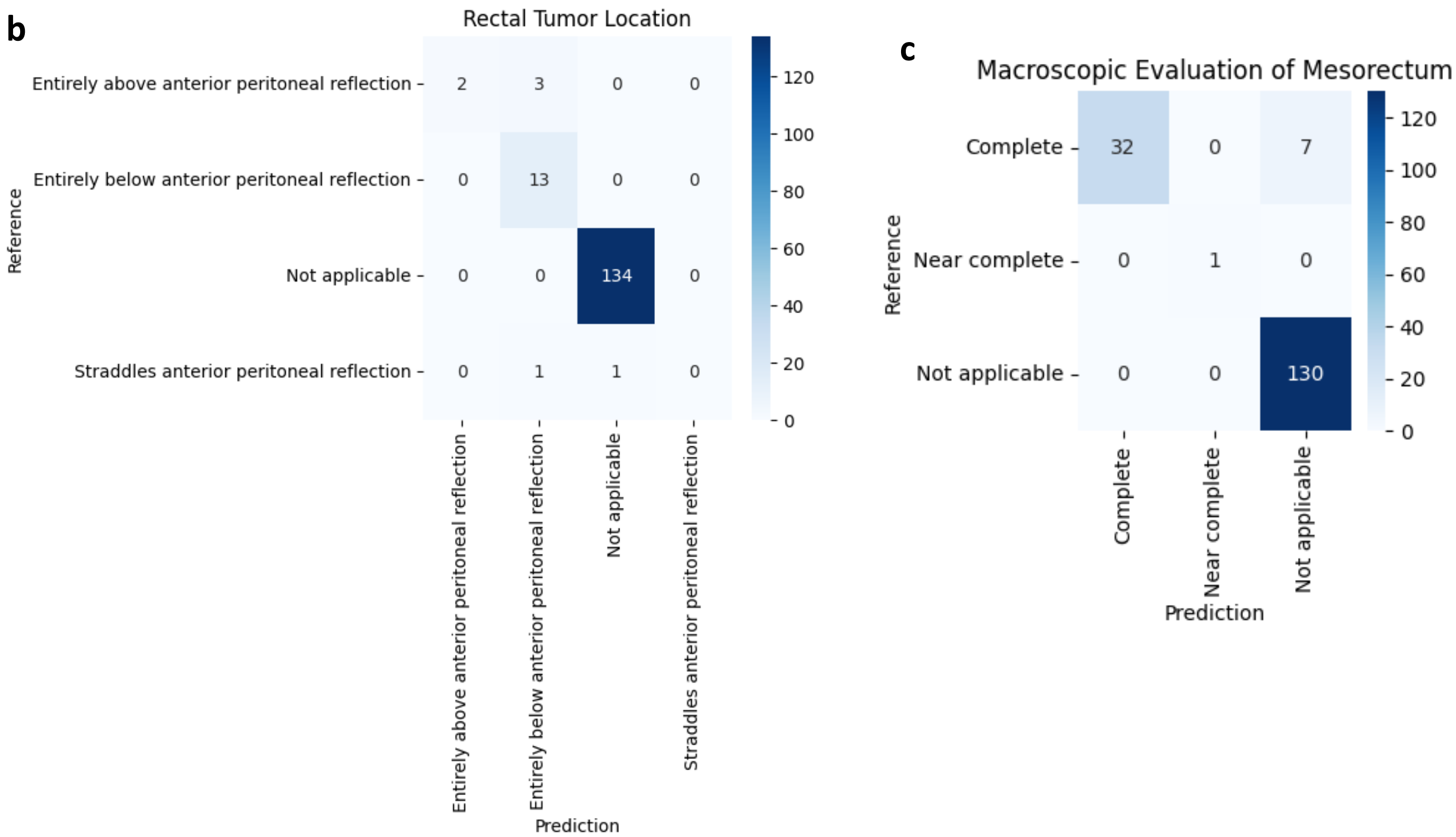

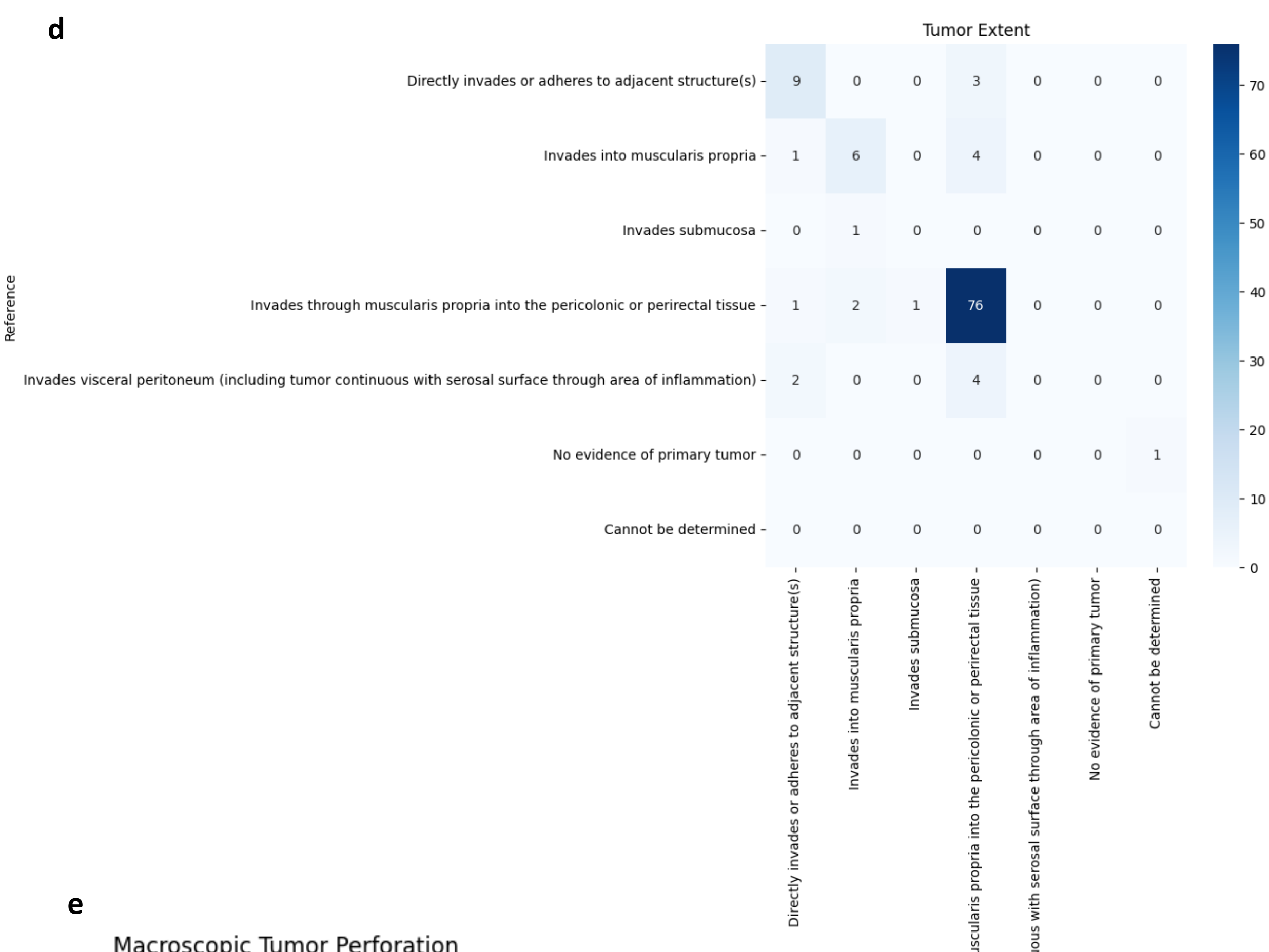

e

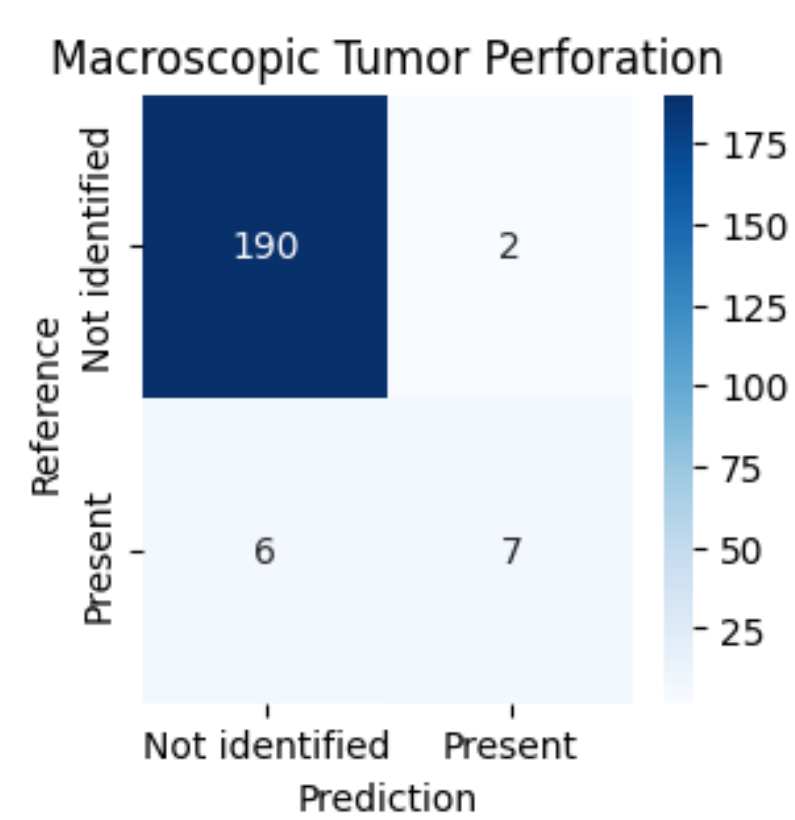

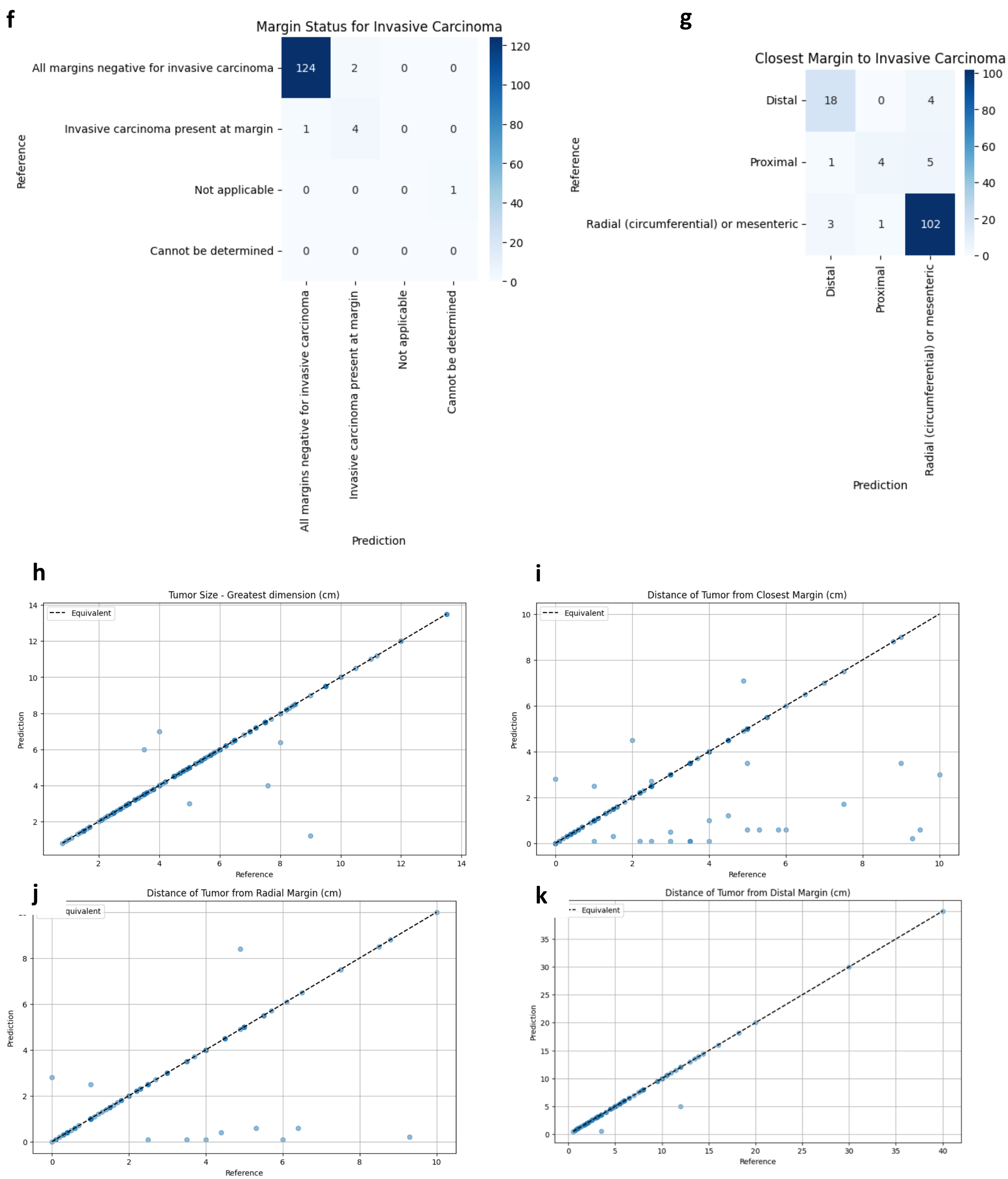

**a**. Confusion matrix of Tumor Site; **b**. Confusion matrix of Rectal Tumor Location; **c**. Confusion matrix of Macroscopic Evaluation of Mesorectum; **d**. Confusion matrix of Tumor Extent; **e**. Confusion matrix of Macroscopic Tumor Perforation; **f**. Confusion matrix of Margin Status; **g**. Confusion matrix of Closest Margin to Invasive Carcinoma; **h**. Tumor Size; **i**. Distance of Tumor from Closest Margin; **j**. Radial Margin; **k**. Distal Margin

## Fig. 4 Analysis of reflections and convergence

**a**

| Predictor/Reflection round | Average F1 (categorical variables) | Average Correct Rate (numeric variables | Cases revised |
|---|---|---|---|
| llama3.3-70b64k_rag03_1by1 | 0.8277 | 0.8058 | 217 |
| Reflection round 1 | 0.9033 | 0.8870 | 195 |
| Reflection round 2 | 0.9094 | 0.8953 | 54 |
| Reflection round 3 | 0.9052 | 0.8953 | 28 |
| Reflection round 4 | 0.9100 | 0.8953 | 27 |
| Reflection round 5 | 0.9052 | 0.8953 | 25 |
| Reflection round 6 | 0.9104 | 0.8953 | 25 |
| Reflection round 7 | 0.9054 | 0.8953 | 25 |
| Reflection round 8 | 0.9104 | 0.8953 | 25 |
| Reflection round 9 | 0.9054 | 0.8953 | 24 |
| Reflection round 10 | 0.9112 | 0.8953 | 24 |
| Reflection round 11 | 0.9061 | 0.8953 | 24 |
| Reflection round 12 | 0.9112 | 0.8953 | 24 |
| Reflection round 13 | 0.9061 | 0.8953 | 24 |
| … | | | |

**b**

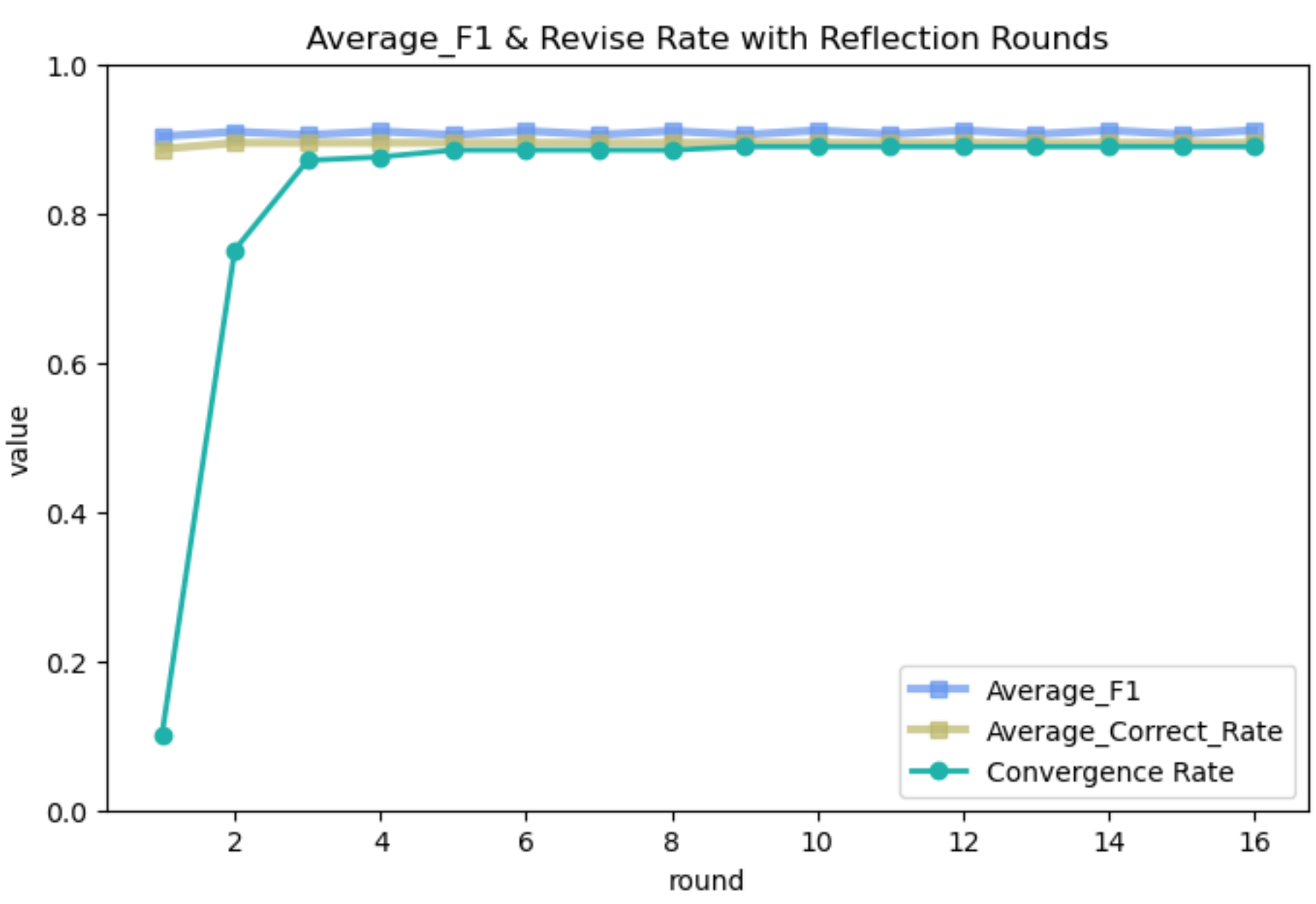

**c**

**Gross Description**

Specimen A received in formalin, labeled with the patient's name and 'sigmoid colon' is a previously partially opened portion of bowel which includes distal sigmoid and proximal rectum. It measures 24.0 cm in length and varies from 2.5 to 4.0 cm in diameter. The proximal resection margin is stapled. The distal margin is open. There is a moderate amount of attached pericolonic adipose tissue. The serosal aspect is focally hemorrhagic and smooth. On opening the bowel the proximal mucosa appears grossly unremarkable. 11 cm distal to the proximal resection margin the mucosa reveals an ill-defined, pink-tan, bosselated area measuring 3.5 cm in the long axis. Just distal to this area there is a circumferential deep ulceration that measures 2.0 cm in the long axis. Also seen within the ulcer bed is gray white tumor tissue which has penetrated the full thickness of the thinned muscularis and appears to have begun to involve the overlying pericolonic adipose tissue. Sectioning through this area reveals a markedly thin muscularis in the ulcer bed and a possible perforation at one point. Just distal to this circumferential ulceration there is a large pocket measuring 3.0 cm in greatest dimension. This pocket contains a stony, hard, green-brown, foreign body consistent with fecalith. It measures 2.5 cm in greatest dimension and has a sharp edge. Distal to this foreign body the mucosa is hyperemic and appears somewhat hypertrophied. Sectioning through this area reveals a thickened mucosa measuring up to 0.8 cm. Distal to this thickened mucosa the remaining 3.0 cm of mucosa appears grossly unremarkable to the distal resection margin. Also received in the same container is a cylindrical wire mesh stent measuring 13.0 cm in length and 2.5 cm in diameter. Sectioning through the pericolonic adipose tissue reveals 19 apparent lymph nodes ranging in size from 0.3 to 1.2 cm.

**Tumor Extent**
Ground Truth: **"**Invades through muscularis propria into the pericolonic or perirectal tissue**"**

**Margin Status**
Ground Truth: **"**All margins negative for invasive carcinoma**"**

**Tumor Extent**: **v0 --> v1 (Reflection Round 9, 11, 13, …)**
v0: Invades through muscularis propria into the pericolonic or perirectal tissue
v1: Directly invades or adheres to adjacent structure(s)
Explanation: "The tumor extent is consistent with direct invasion of adjacent structures."
['the tumor has penetrated the full thickness of the thinned muscularis and appears to have begun to involve the overlying pericolonic adipose tissue.']

**Margin Status**: **v0 unchanged (Reflection Round 9)**
v0: Invasive carcinoma present at margin
Explanation: "The margin status for invasive carcinoma is identified as present at margin based on the gross description.."
['Just distal to this area there is a circumferential deep ulceration that measures 2.0 cm in the long axis.']

**Tumor Extent**: **v0 --> v1 (Reflection Round 10, 12, 14, … )**
v0: Directly invades or adheres to adjacent structure(s)
v1: Invades through muscularis propria into the pericolonic or perirectal tissue
Explanation: "The tumor extent is identified as invades through muscularis propria into the pericolonic or perirectal tissue based on the gross description."
['the tumor has penetrated the full thickness of the thinned muscularis and appears to have begun to involve the overlying pericolonic adipose tissue.']

**Margin Status**: **v0 --> v1. (Reflection Round 10, 11, 12, 13, 14, … )**
v0: Invasive carcinoma present at margin
v1: All margins negative for invasive carcinoma
Explanation: "The margin status for invasive carcinoma is identified as all margins negative for invasive carcinoma based on the gross description.."
['the tumor has penetrated the full thickness of the thinned muscularis and appears to have begun to involve the overlying pericolonic adipose tissue.', '11 cm distal to the proximal resection margin']

**Analysis:** LLM confused "pericolonic adipose tissue" with "adjacent structure(s)", thus oscillating between two related values of tumor extent.
LLM understood "just distal" as margin positive for tumor and treated "a circumferential deep ulceration" as the tumor.

**a.** The table shows the average F1 score for all categorical synoptic data estimation, average correct rate, and the number of cases revised (where the rest of cases remain unchanged, i.e. converged in reflection) for all numeric synoptic data estimation in the base estimation (one-by-one query with RAG) and each round of reflective reasoning. **b.** The curves illustrate the changes of Average F1 score, the average correct rate, and the convergence rate with the rounds of reflective reasoning. **c.** An example, case R0130 (which is one of 24 cases not converged in reflection), showcases the reflection process revises previous estimation and oscillates between two standard values of tumor extent.

## Table 2. Performance comparison with different alternatives

<table>
<tr><th>Predictor<br>(with llama3.3-70b)</th><th>Average<br>F1</th><th>Average<br>Correct Rate</th><th>Tokens<br>(input; output)</th><th>Computing<br>time</th></tr>
<tr><td>1. One-by-one QA without KB</td><td>0.7577</td><td>0.8313</td><td>2,647,793; 406,525</td><td>5.79 hours</td></tr>
<tr><td>2. One-by-one QA with RAG KB</td><td>0.8277</td><td>0.8058</td><td>5,810,655; 481,204</td><td>7.16 hours</td></tr>
<tr><td>3. Reflection on one-by-one QA with RAG KB<br>(Round 10, converged, best)</td><td>0.9112</td><td>0.8953</td><td rowspan="2">6,838,020; 1,082,968</td><td rowspan="2">18.51 hours<br>(16 rounds of reflections)</td></tr>
<tr><td>4. Reflection on one-by-one QA with RAG KB<br>(Round 11, worst in converged)</td><td>0.9061</td><td>0.8953</td></tr>
</table>

# Fig. 5 Error analysis on tumor site estimation

**a**

CAP Approved

ColoRectal_4.2.0.2.REL_CAPCP

**Gross Description**

…

It consists of a 8.0 cm length by up to 3.3 cm in diameter segment of terminal ileum, attached to a 18.5 cm in length by up to 7.5 cm circumference segment of large bowel.

…

The specimen is opened to reveal a 4.0 cm length by 8.5 cm in circumference endophytic mass with raised borders (corresponding to the previously mentioned serosa umbilication) located within the proximal aspect of the ascending colon, 8.5 cm from the proximal margin, 12.0 cm from the distal margin and 6.2 cm from the nearest deep/mesenteric margin.

The mass appears to involve the ileocecal valve.

...

**Llama 3.3-70b One-by-one with RAG output:**

Tumor Site: Ascending colon

Explanation: "The tumor site is inferred as the ascending colon because the gross description mentions a mass located within the proximal aspect of the ascending colon. Although the mass involves the ileocecal valve, which is the border between the small intestine and the large intestine (at the junction of the terminal ileum and cecum), the primary location of the tumor is described as being in the ascending colon."

['a 4.0 cm length by 8.5 cm in circumference endophytic mass with raised borders located within the proximal aspect of the ascending colon', 'the mass appears to involve the ileocecal valve']

**Reflections output:**

Tumor Site: Ascending colon

Explanation: "The gross description explicitly states that the tumor is located in the proximal aspect of the ascending colon.."

['A 4.0 cm length by 8.5 cm in circumference endophytic mass with raised borders... located within the proximal aspect of the ascending colon']

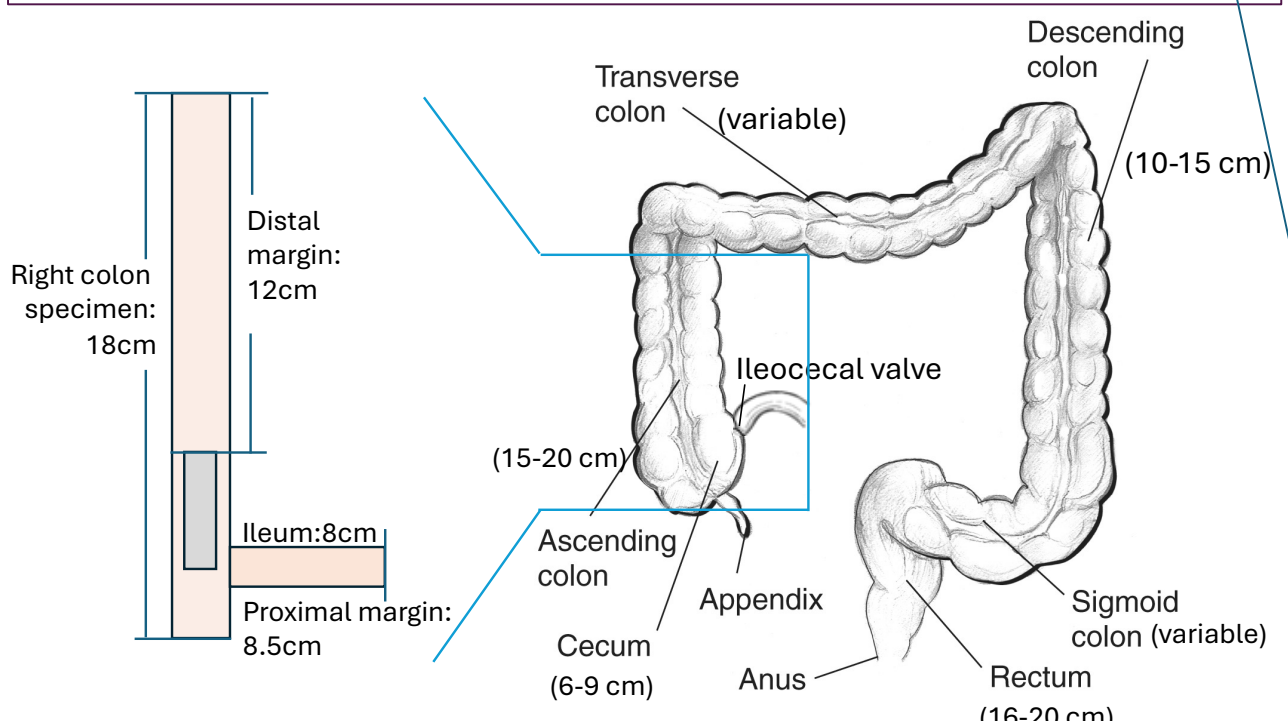

**Tumor Site (Ground Truth): Cecum**

CAP Approved

**Error Analysis**:

The gross description about tumor site is a bit complex. The LLM focused on partial information and missed key part of "8.5 cm from the proximal margin, 12.0 cm from the distal margin". Probably phrase "located **within** the proximal aspect of the ascending colon" and 8.5 cm proximal margin mislead LLM to think that the tumor involves ascending colon. LLM reflections failed to correct the error for the same mistakes. The task of inferring tumor site for this case is more challenge for involving proper understanding of anatomic structure.

ColoRectal_4.2.0.2.REL_CAPCP

**b**

**Gross Description**

…

Cecum and ascending bowel: 27.0 cm in length and dilated up to 7.5 cm in diameter within the proximal cecum

…

Tumor Description: Measuring approximately 5.5 cm from the distal resection margin and greater than 15.0 cm from the proximal resection margin is a fully circumferential, centrally ulcerated, pink-tan mass with raised serpiginous borders. The mass is 7.5 x 5.5 cm. Grossly, the mass invades the muscularis and the pericolonic adipose tissue. …

Within the cecum, there is a 1.5 cm perforation with surrounding induration of mesentery and fibrinopurulent kinking. The abscess area is 11.5 x 9.5 cm and involves the appendix. ...

**Llama 3.3-70b One-by-one with RAG output:**

Tumor Site: Cecum

Explanation: "The tumor description mentions that the mass is located in the cecum, with a perforation also present in the cecum. The dimensions and location of the mass suggest that it is primarily situated in the cecum."

['Measuring approximately 5.5 cm from the distal resection margin and greater than 15.0 cm from the proximal resection margin is a fully circumferential, centrally ulcerated, pink-tan mass with raised serpiginous borders. The mass is 7.5 x 5.5 cm.', 'Within the cecum, there is a 1.5 cm perforation with surrounding induration of mesentery and fibrinopurulent kinking.']

**Reflections output:**

Tumor Site: Cecum

Explanation: "The tumor is described as being in the cecum based on its location relative to the resection margins."

['A fully circumferential, centrally ulcerated, pink-tan mass with raised serpiginous borders.', 'Measuring approximately 5.5 cm from the distal resection margin and greater than 15.0 cm from the proximal resection margin']

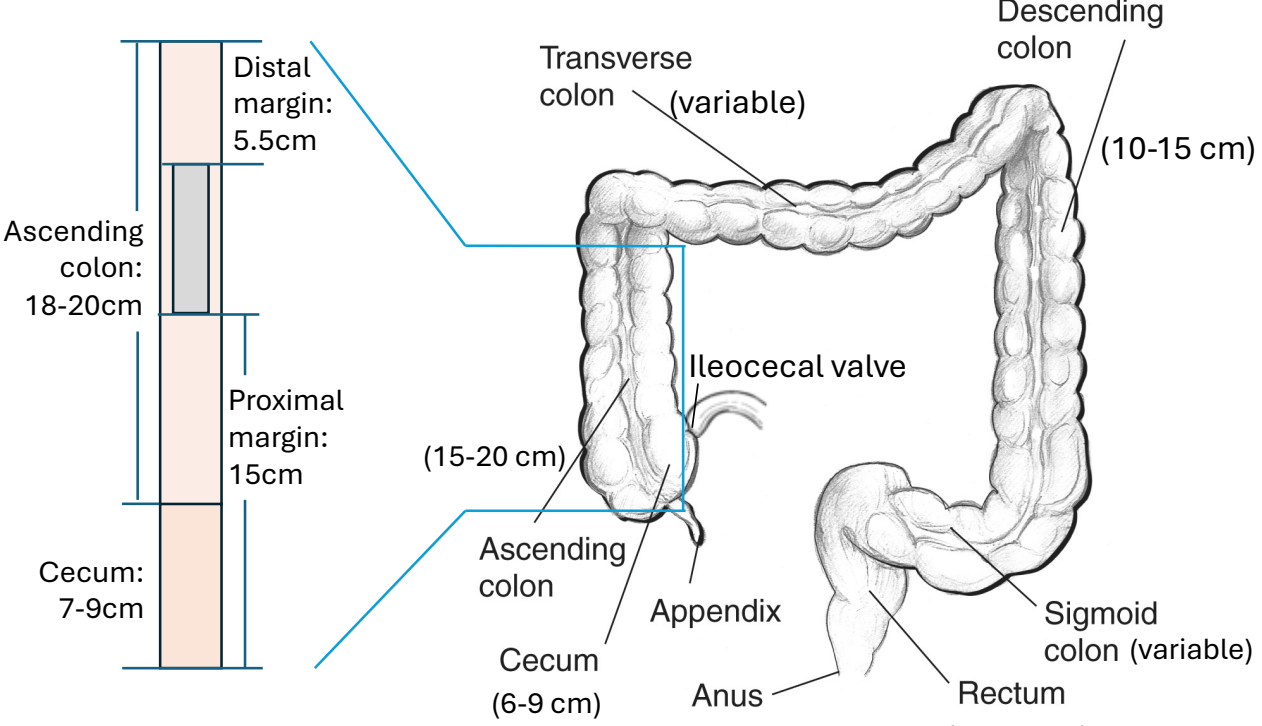

Tumor Site (Ground Truth): Ascending colon

**Error Analysis**:

The gross description about tumor is not explicit and needs to be inferred. The LLM made mistakes in reasoning about tumor site, although picked relevant text. The mistakes may come from the weakness in geometric and numeric reasoning or misled by text about cecum. LLM reflections failed to correct the error.

Error analysis about tumor site estimation for two cases: **a.** case R014, **b.** case R044.

| Site | Relationship to Peritoneum (see Note J) | Dimensions (approximate) |
|---|---|---|
| Cecum | Entirely covered by peritoneum | 6-9 cm |
| Ascending colon | Retroperitoneal; posterior surface lacks peritoneal | 15-20 cm |
| Transverse colon | Intraperitoneal; has mesentery | Variable |
| Descending colon | Retroperitoneal; posterior surface lacks peritoneal | 10-15 cm |
| | peritoneum (serosa) | |
| Sigmoid colon | Intraperitoneal; has mesentery | Variable |
| Rectum | Upper third covered by peritoneum on anterior and lateral | 16-20 cm |
| | anterior surface; lower third has no peritoneal covering | |

## Fig. 6 Case analysis of inferring margins with case R0227

**a**

**Gross Description**

Specimen A, received in formalin labeled with the patient's name designated 'terminal ileum, appendix, right colon', is an ileum with attached cecum, partial right colon and appendix. The terminal ileum measures 42.0 cm in length x 3.0 cm in diameter. The right colon and cecum measure 17.0 cm in length x up to 4.0 cm in diameter. The ileal mucosa demonstrates flattened mucosa with areas of ulceration ranging in size up to 3.3 cm in greatest dimension. The closest area of ulceration is located 10.0 cm from the proximal margin. The cecum demonstrates flattened mucosa with a thickened wall. The right colon demonstrates an area measuring approximately 7.5 cm in length, which demonstrates a thickened wall, cobblestoned mucosa, and an area of stricture, which extends to the distal margin. The serosa is inked black. No discrete masses are identified. Multiple lymph nodes are identified, some of which are grossly positive, and are completely submitted. Further sectioning through the area of ulceration reveals an area of invasion into the surrounding adipose tissue measuring approximately 1.0 x 0.7 x 0.5 cm. The area of ulceration encroaching into adipose tissue is located 2.5 cm away from the nearest mesenteric margin.

The appendix measures 5.0 cm in length x 0.6 cm in diameter, and is unremarkable.

Section Key:
A1 - proximal ileal margin
A2-A4 - representative sections of ileum
A5 - ileocecal valve
A6, A7 - representative sections of cecum and right colon
…
A23-A26- additional sections of tumor to serosal surface

**One-by-one with RAG output:**

Tumor Site: Cecum

Explanation: "The tumor site is inferred to be the cecum based on the description of the cecum demonstrating flattened mucosa with a thickened wall, and the presence of the ileocecal valve (A5) which is a landmark for the junction between the ileum and the cecum. Additionally, sections A6 and A7 are representative of the cecum and right colon, indicating that the tumor involves the cecum."

**Reflections output:**

Tumor Site: Cecum

Explanation: "The tumor site is identified as the cecum based on the description of the specimen."

['The cecum demonstrates flattened mucosa with a thickened wall.']

**Tumor Site - Ground Truth: Ascending colon**

**Error Analysis**: The gross description states multiple lesions. The LLM misjudged major lesion and picked wrong pieces of text rather than right ones as evidence, leading to wrong conclusion in the initial estimation. The LLM reflections failed to correct the error for the same wrong perspective.

**b**

**Gross Description**

Specimen A, received in formalin labeled with the patient's name designated 'terminal ileum, appendix, right colon', is an ileum with attached cecum, partial right colon and appendix. The terminal ileum measures 42.0 cm in length x 3.0 cm in diameter. The right colon and cecum measure 17.0 cm in length x up to 4.0 cm in diameter. The ileal mucosa demonstrates flattened mucosa with areas of ulceration ranging in size up to 3.3 cm in greatest dimension. The closest area of ulceration is located 10.0 cm from the proximal margin. The cecum demonstrates flattened mucosa with a thickened wall. The right colon demonstrates an area measuring approximately 7.5 cm in length, which demonstrates a thickened wall, cobblestoned mucosa, and an area of stricture, which extends to the distal margin. The serosa is inked black. No discrete masses are identified. Multiple lymph nodes are identified, some of which are grossly positive, and are completely submitted. Further sectioning through the area of ulceration reveals an area of invasion into the surrounding adipose tissue measuring approximately 1.0 x 0.7 x 0.5 cm. The area of ulceration encroaching into adipose tissue is located 2.5 cm away from the nearest mesenteric margin.

The appendix measures 5.0 cm in length x 0.6 cm in diameter, and is unremarkable.

Section Key:
A1 - proximal ileal margin
A2-A4 - representative sections of ileum
A5 - ileocecal valve
A6, A7 - representative sections of cecum and right colon
…
A23-A26- additional sections of tumor to serosal surface

**One-by-one with RAG output**:

Distance of Tumor from Radial Margin (cm): 0.5

Explanation: "The distance from the tumor to the radial margin is inferred based on the description of the area of ulceration encroaching into adipose tissue, which is located **2.5** cm away from the nearest mesenteric margin. However, since the tumor invades into the surrounding adipose tissue with a dimension of **0.5** cm, it can be inferred that the distance from the tumor to the radial margin is approximately 0.5 cm or less, given the context of 'invasion into the surrounding adipose tissue'."

**Reflections output:**

Distance of Tumor from Radial Margin (cm): 2.5

Explanation: "The distance from the tumor to the radial margin is estimated based on the description."

['The area of ulceration encroaching into adipose tissue is located 2.5 cm away from the nearest mesenteric margin.']

**Distance of Tumor from Radial Margin (cm) - Ground Truth: 2.5**

**Error Analysis**: LLM collected relevant evidence but misunderstood and confused concept of radial margin with lesion dimensions, leading to wrong conclusion. LLM reflections correctly fixed the error.

**c**

**Gross Description**

Specimen A, received in formalin labeled with the patient's name designated 'terminal ileum, appendix, right colon', is an ileum with attached cecum, partial right colon and appendix. The terminal ileum measures 42.0 cm in length x 3.0 cm in diameter. The right colon and cecum measure 17.0 cm in length x up to 4.0 cm in diameter. The ileal mucosa demonstrates flattened mucosa with areas of ulceration ranging in size up to 3.3 cm in greatest dimension. The closest area of ulceration is located 10.0 cm from the proximal margin. The cecum demonstrates flattened mucosa with a thickened wall. The right colon demonstrates an area measuring approximately 7.5 cm in length, which demonstrates a thickened wall, cobblestoned mucosa, and an area of stricture, which extends to the distal margin. The serosa is inked black. No discrete masses are identified. Multiple lymph nodes are identified, some of which are grossly positive, and are completely submitted. Further sectioning through the area of ulceration reveals an area of invasion into the surrounding adipose tissue measuring approximately 1.0 x 0.7 x 0.5 cm. The area of ulceration encroaching into adipose tissue is located 2.5 cm away from the nearest mesenteric margin.

The appendix measures 5.0 cm in length x 0.6 cm in diameter, and is unremarkable.

Section Key:

A1 - proximal ileal margin

A2-A4 - representative sections of ileum

A5 - ileocecal valve

A6, A7 - representative sections of cecum and right colon

…

A23-A26- additional sections of tumor to serosal surface

**One-by-one with RAG output:**

Distance of Tumor from Distal Margin (cm) : 0

Explanation: "The tumor extends to the distal margin, indicating that the distance between the tumor and the distal margin is 0 cm. This is inferred from the phrase 'which extends to the distal margin' in the gross description."

**Reflections output:**

Distance of Tumor from Distal Margin (cm) : 0

Explanation: "The tumor is described as extending to the distal margin."

['The right colon demonstrates an area measuring approximately 7.5 cm in length, which demonstrates a thickened wall, cobblestoned mucosa, and an area of stricture, which extends to the distal margin.']

**Distance of Tumor from Distal Margin (cm) - Ground Truth**: 0

**Case analysis**: Both the base LLM predictor and the LLM reflections correctly inferred the distal margin.

**d**

**Gross Description**

Specimen A, received in formalin labeled with the patient's name designated 'terminal ileum, appendix, right colon', is an ileum with attached cecum, partial right colon and appendix. The terminal ileum measures 42.0 cm in length x 3.0 cm in diameter. The right colon and cecum measure 17.0 cm in length x up to 4.0 cm in diameter. The ileal mucosa demonstrates flattened mucosa with areas of ulceration ranging in size up to 3.3 cm in greatest dimension. The closest area of ulceration is located 10.0 cm from the proximal margin. The cecum demonstrates flattened mucosa with a thickened wall. The right colon demonstrates an area measuring approximately 7.5 cm in length, which demonstrates a thickened wall, cobblestoned mucosa, and an area of stricture, which extends to the distal margin. The serosa is inked black. No discrete masses are identified. Multiple lymph nodes are identified, some of which are grossly positive, and are completely submitted. Further sectioning through the area of ulceration reveals an area of invasion into the surrounding adipose tissue measuring approximately 1.0 x 0.7 x 0.5 cm. The area of ulceration encroaching into adipose tissue is located 2.5 cm away from the nearest mesenteric margin.

The appendix measures 5.0 cm in length x 0.6 cm in diameter, and is unremarkable.

Section Key:

A1 - proximal ileal margin

A2-A4 - representative sections of ileum

A5 - ileocecal valve

A6, A7 - representative sections of cecum and right colon

…

A23-A26- additional sections of tumor to serosal surface

**One-by-one with RAG output:**

Distance of Tumor from Closest Margin (cm): 0.5

Explanation: "The closest margin to the tumor is the mesenteric margin, which is **2.5** cm away from the area of ulceration encroaching into adipose tissue. However, since the tumor has invaded into the surrounding adipose tissue with a dimension of 0.5 cm, we can infer that the distance from the tumor to the closest margin is less than or equal to **0.5** cm, given the context of radial margin assessment."

**Reflections output:**

Distance of Tumor from Closest Margin (cm): 0

Explanation: "The tumor is described as extending to the closest margin."

['The right colon demonstrates an area measuring approximately 7.5 cm in length, which demonstrates a thickened wall, cobblestoned mucosa, and an area of stricture, which extends to the distal margin.']

**Distance of Tumor from Closest Margin (cm) - Ground Truth**: 0

**Error Analysis**: Probably due to the mistake made on radial margin, the LLM made the same mistake again; and even worse, the LLM failed to take into account of distal margin of zero, reflecting insufficient understanding of what is closest margin. The LLM reflections correctly fixed the error.

**e**

**Gross Description**

Specimen A, received in formalin labeled with the patient's name designated 'terminal ileum, appendix, right colon', is an ileum with attached cecum, partial right colon and appendix. The terminal ileum measures 42.0 cm in length x 3.0 cm in diameter. The right colon and cecum measure 17.0 cm in length x up to 4.0 cm in diameter. The ileal mucosa demonstrates flattened mucosa with areas of ulceration ranging in size up to 3.3 cm in greatest dimension. The closest area of ulceration is located 10.0 cm from the proximal margin. The cecum demonstrates flattened mucosa with a thickened wall. The right colon demonstrates an area measuring approximately 7.5 cm in length, which demonstrates a thickened wall, cobblestoned mucosa, and an area of stricture, which extends to the distal margin. The serosa is inked black. No discrete masses are identified. Multiple lymph nodes are identified, some of which are grossly positive, and are completely submitted. Further sectioning through the area of ulceration reveals an area of invasion into the surrounding adipose tissue measuring approximately 1.0 x 0.7 x 0.5 cm. The area of ulceration encroaching into adipose tissue is located 2.5 cm away from the nearest mesenteric margin.

The appendix measures 5.0 cm in length x 0.6 cm in diameter, and is unremarkable.

Section Key:
A1 - proximal ileal margin
A2-A4 - representative sections of ileum
A5 - ileocecal valve
A6, A7 - representative sections of cecum and right colon
…
A23-A26- additional sections of tumor to serosal surface

**One-by-one with RAG output:**

Margin Status: "All margins negative for invasive carcinoma"

Explanation: "The gross description indicates that the closest area of ulceration is 10.0 cm away from the proximal margin, and the area of invasion into the surrounding adipose tissue is 2.5 cm away from the nearest mesenteric margin. This suggests that the tumor is not present at the margins. Additionally, the serosa is inked black, which helps to identify the margin. There is no indication of tumor involvement at the distal margin, as the stricture extends to but does not involve the distal margin."

**Reflections output:**

Margin Status: "Invasive carcinoma present at margin"

Explanation: "The margin status is estimated as invasive carcinoma present at the margin based on the description."

['The right colon demonstrates an area measuring approximately 7.5 cm in length, which demonstrates a thickened wall, cobblestoned mucosa, and an area of stricture, which extends to the distal margin.']

**Margin Status - Ground Truth**: "Invasive carcinoma present at margin"

**Error Analysis**: In initial estimation, the LLM made wrong judgment, although it considered all relevant evidence, including somewhat ambiguous description of "extends to the distal margin", which the LLM argued that "extends to but does not involve the distal margin." The LLM reflections correctly fixed the error.

This figure demonstrates the effects of reflective reasoning through error analysis with case R0227 as an example, where in the initial estimation, the LLM (Llama3.3-70b) frequently made mistakes due to confusing with key concepts for improper understanding and biased in judgment with partial or secondary evidence; fortunately, the LLM could identify and fix most of the errors in its reflective reasoning. **a**. Inferring Tumor Site, the LLM's initial estimation was incorrect due to wrong judgment with wrong evidence; the LLM reflections failed to correct for the same wrong perspective. **b**. Inferring Radial Margin, the LLM's initial estimation was wrong due to wrong with concepts; the LLM reflections correct the error. **c**. Inferring Distal Margin, the LLM made correct judgment in both initial estimation and reflections. **d**. Inferring Closest Margin, the LLM's initial estimation made the same mistake as in b and missed key information of zero distal margin; the LLM reflections successfully fixed the problem. **e**. Inferring Margin Status, the LLM made mistake due to somewhat ambiguous gross description; the LLM reflections correctly fix the error.

# Supplementary materials

## Fig. S1 Prompt used for one-by-one query

```
f"""You are an AI Assistant that follows instructions extremely well. You work as a
pathologist assistant helping to extract and infer structured pathology synoptic data from
clinic notes in the field of Colon and Rectum Carcinoma.

# **Tasks**

Based on the facts given in **Gross Description** section, infer the standard value for
synoptic data element "{var}", by selecting one value from the following list of possible
standard values

{values}.

Output your answer in JSON format as follows.

{
"{var}": "<Inferred value>",
"{var} - Description": "<List each piece of original text extracted from the gross
description related to the variable.>",
"{var} - Explanation": "<Explanation of inference>",
"{var} - Belief Degree": "<A real number between 0 and 1, indicating your degree of belief
in the inferred value.>"
}

Only output the queried JSON data, nothing else.

# **Gross Description**
<Gross Description>
{gross_description}
</Gross Description>
"""
```

Fig. S2 Prompt used for reflections

```
f"""You are an AI Assistant that follows instructions extremely well. You work as a pathologist assistant helping to extract and infer structured pathology synoptic data from clinic notes in the field of Colon and Rectum Carcinoma.

# **Instructions**

Your overarching goal is to estimate the synoptic data elements from the input **Gross Description** of a colorectal cancer pathology report.

Before you start, review the following data, given in the **Input Data** section:
1. **Gross Description**
2. **Relevant Knowledge**
3. **Synoptic Data Template**
4. **Synoptic Data Estimation** (initial synoptic value assignments)

Synoptic data elements are interdependent to each other. Consider the associations among them. The values assigned to synoptic data variables may become inconsistent, when the combinations of the values cannot be true at the same time. There are two types of inconsistencies as follows:

* Type 1 - Inconsistencies with facts: Inconsistencies between the value of a variable and the facts given in **Gross Description**. This type of inconsistencies occur, when the related text in **Gross Description** cannot be interpreted as a variable value given in **Synoptic Data Estimation**. Note that any variable value given in **Synoptic Data Estimation** should be a standard value listed in **Synoptic Data Template**; however, the terms used in **Gross Description** are free text without constraints, therefore, the standard value of a variable may be different to the corresponding gross description text; however the text in **Gross Description** can imply or be interpreted as that standard value. This type of term difference is **NOT** inconsistency.

* Type 2 - Inconsistencies among the values of variables: Inconsistencies among values of interdependent variables, that is, the value of a variable and the value of another variable cannot be true at the same time.

{inconsistence_examples}
```

## **Tasks**

* Review the initial synoptic value assignments given in **Synoptic Data Estimation** to identify inconsistencies. Pay attention that a difference between (a) the non-standard value description in **Gross Description** and (b) the best-matching standard value listed in **Synoptic Data Template** should not be regarded as an inconsistence. For example, *Tumor Site* was assessed as "Ascending colon" but tumor was described in "right colon" in gross description; this case **is not an inconsistence** and **don't change initial value of "Ascending colon" to "Right colon", unless you are very certain that the tumor is indeed at cecum.

* Revise the initial synoptic value assignments given in **Synoptic Data Estimation** to minimize the identified inconsistencies, that is, (1) to make synoptic value assignments consistent with facts, and use the standard values listed in **Synoptic Data Template**; (2) to make synoptic value assignments consistent with each other.

* Hints:
(1) If the gross description says tumor extending to an adjacent organ outside colon or rectum, the consistent value of "Tumor Extent" should be "Directly invades or adheres to adjacent structure(s)".
(2) Discern the meanings of different standard values of "Tumor Extent", particularly the following ones,
"Directly invades or adheres to adjacent structure(s)" means that the tumor extends to other organs outside of colon or rectum;
"Invades visceral peritoneum (including tumor continuous with serosal surface through area of inflammation)" means that tumor reaches the outmost of colon and rectum;
"Invades through muscularis propria into the pericolonic or perirectal tissue" means that tumor has penertrated through muscularis propria but still within the boundary of colorectal structures;
"Invades into muscularis propria" means that tumor is still within muscularis propria.
(3) "Invades through muscularis propria into the pericolonic or perirectal tissue" is quite different from "Invades into muscularis propria". Pay close attention to which case the gross description states, and avoid confusing "invades through" and "invades into".

## **Requirements**

### **JSON Output**
Output all of your answers to all synoptic data element together in a flatten JSON data structure, as follows.

{
"<var_1>": "<for variable_1, inferred a proper standard value, listed in **Synoptic Data Template**>",

"<var_1> - Description": "<A list of the selected pieces of **original text** extracted from the gross description, which serve as supporting evidence for the estimation of variable_1>",
"<var_1> - Explanation": "<Explanation of inference for variable_1, including why to revise the value, if you revised.>",
"<var_1> - Belief Degree": "<A real number between 0 and 1, indicating your degree of belief in the inferred value.>"
...
"<var_n>": "<for variable_n, inferred a proper standard value, listed in **Synoptic Data Template**>",
"<var_n> - Description": "<A list of the selected pieces of **original text** extracted from the gross description, which serve as supporting evidence for the estimation of variable_n>",
"<var_n> - Explanation": "<Explanation of inference for variable_n, including why to revise the value, if you revised.>",
"<var_n> - Belief Degree": "<A real number between 0 and 1, indicating your degree of belief in the inferred value.>"
}

Don't output anything else beyond the above JSON data.

### **Review and revise**
* Review the initial synoptic value assignments given in **Synoptic Data Estimation** carefully.
* Make revisions to initial values, only if you are sure they are inconsistent.
* Make revisions, only if the revisions make overall inconsistence decrease.

### **Standard Synoptic Values**
When you review and revise, **make sure to use a standard value** listed in **Synoptic Data Template**.
When dealing with the non-standard description of the facts presented in **Gross Description**, select a best suitable standard value listed in **Synoptic Data Template**.

### **Discern Special Standard Values**
Discern the difference between values "Not applicable", "Cannot be determined", and "Not specified".
* "Not applicable" means that the synoptic variable is not applicable to the case, e.g. if tumor is not in rectum, then variables *Macroscopic Evaluation of Mesorectum* and *Rectal Tumor Location* become "Not applicable".
* "Cannot be determined" means that there is not enough information to estimate a value for a variable.
* "Not specified" means that there is no information about the variable at all.

# **Input Data**

1. **Gross Description**
<Gross Description>
{gross_description}
</Gross Description>

2. **Relevant Knowledge**
<Relevant Knowledge>
{knowledge_packages}
</Relevant Knowledge>

3. **Synoptic Data Template**
<Synoptic Data Template>
The Synoptic Data Template is in JSON format, where a key is a synoptic data element name (variable), and the key's value is a list of the possible standard values of that synoptic data element.

{CAP_template}
</Synoptic Data Template>

4. **Synoptic Data Estimation** (initial synoptic value assignments)
<Synoptic Data Estimation>
The estimated synoptic data values are provided in JSON format, where a key is a synoptic data element name (or variable), and the value to the key is the synoptic data element value.

{synoptic_data_estimation}
</Synoptic Data Estimation>
"""

Fig. S3 Error analysis of case R0016

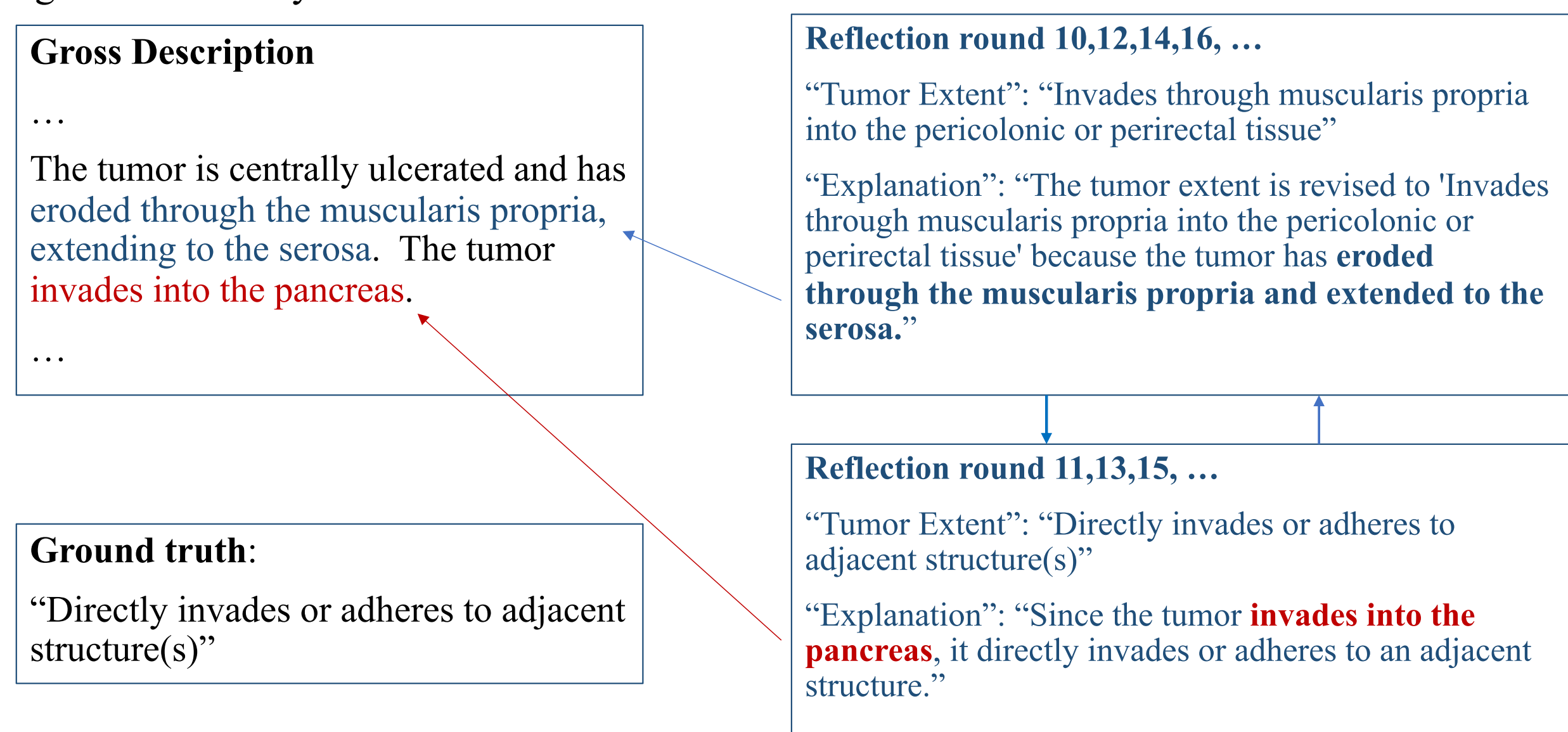

In this case, LLM focused its attention on partial information rather than the whole. In reflection round 10, LLM correctly identified the first part of relevant information "eroded through the muscularis propria, extending to the serosa", but neglected the second part "invades into the pancreas", which is more critical. In round 11, LLM accurately caught the key information, and predict tumor extent correctly. However, in the next round, LLM flipped back to the wrong end.

## Fig. S3 Deep reflective reasoning for lung TNM staging

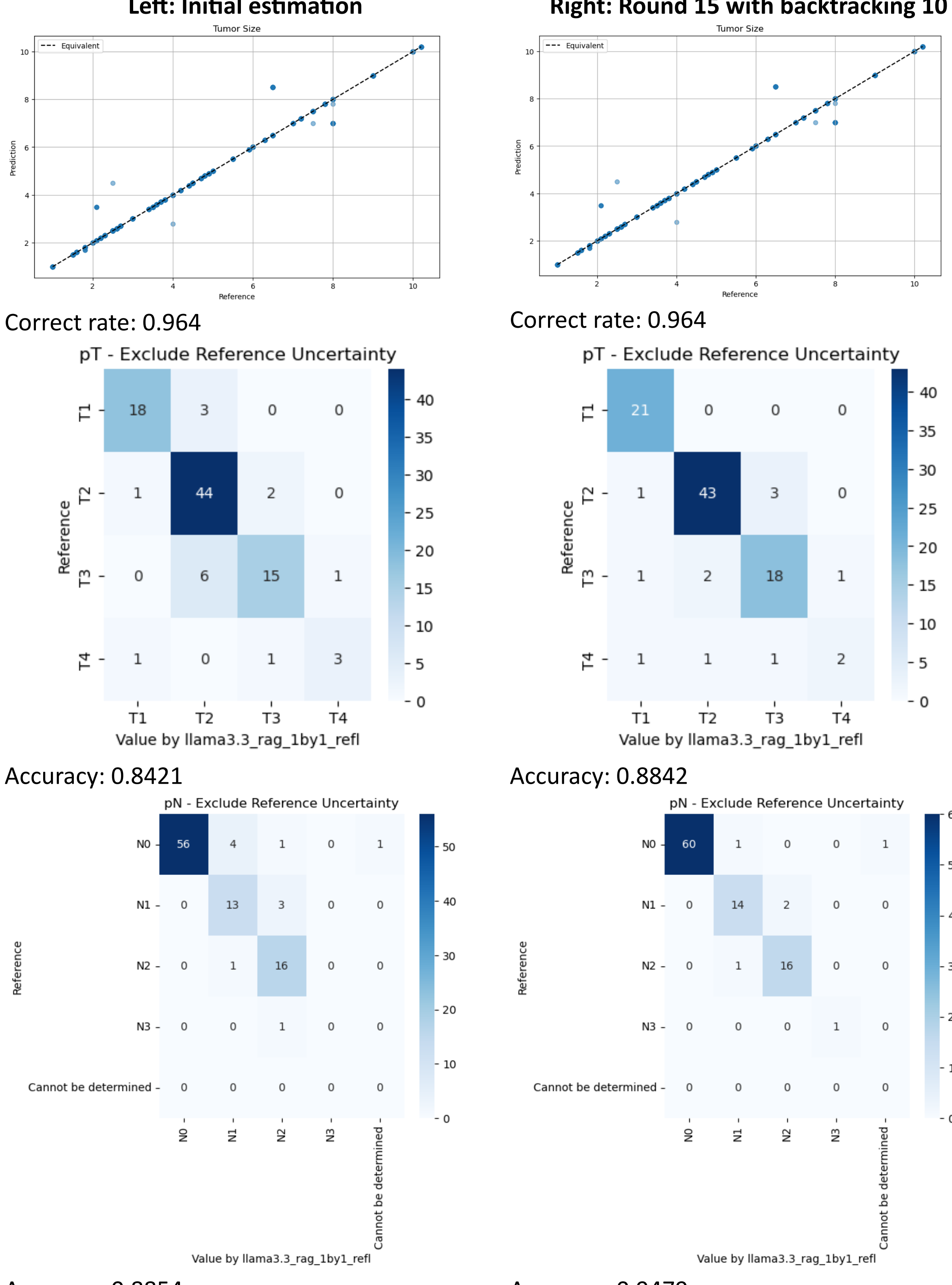

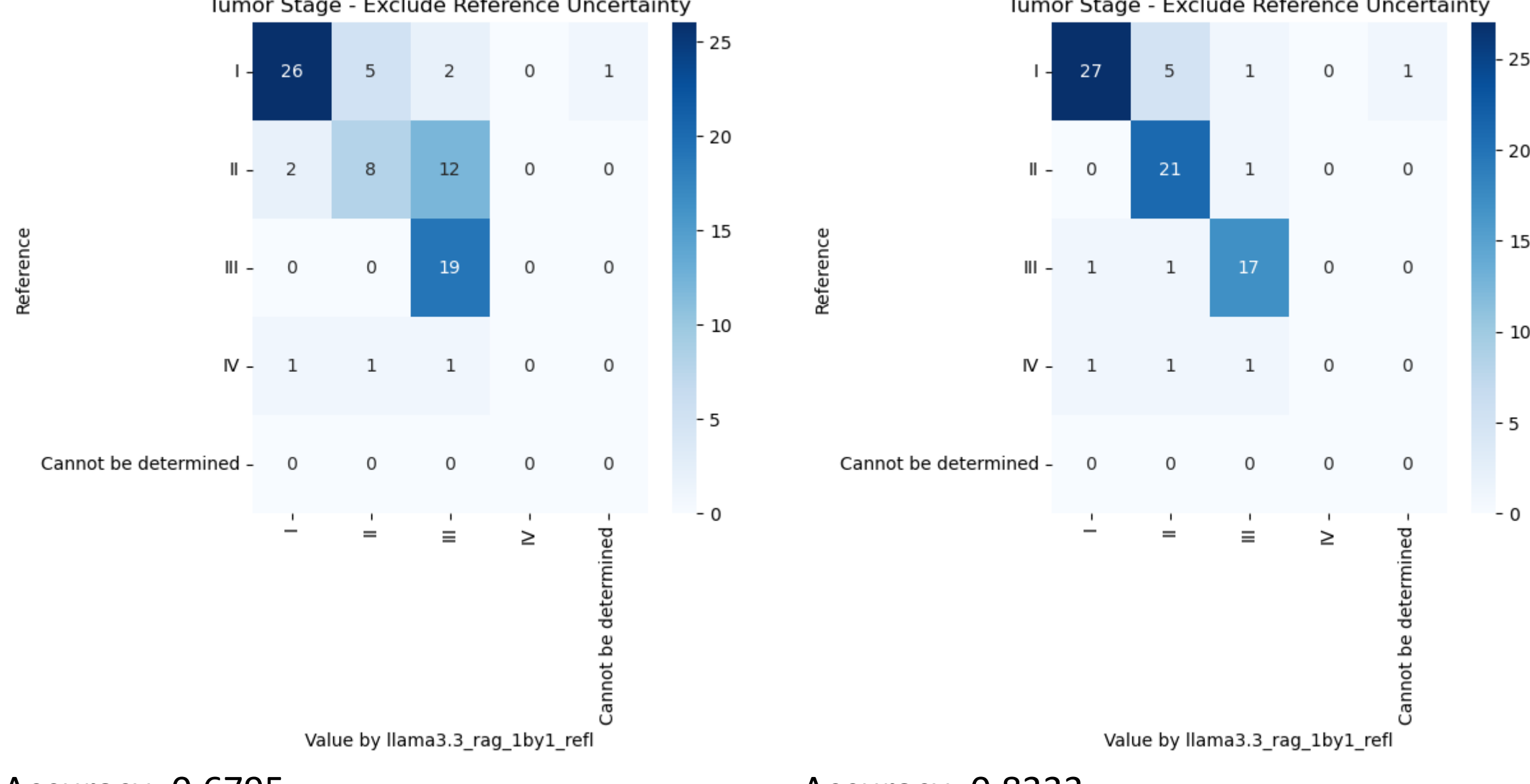

Accuracy: 0.6795

Accuracy: 0.8333

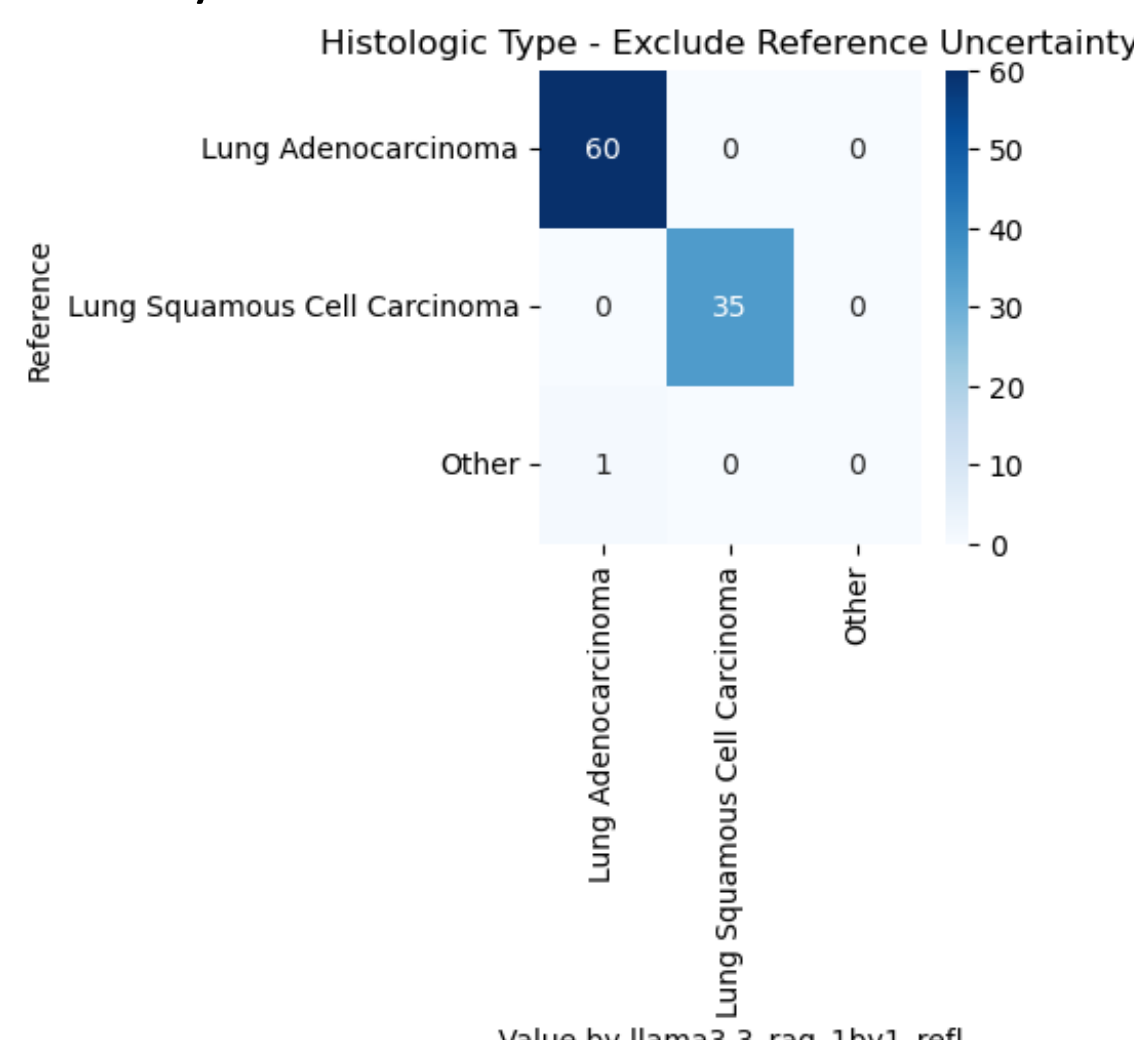

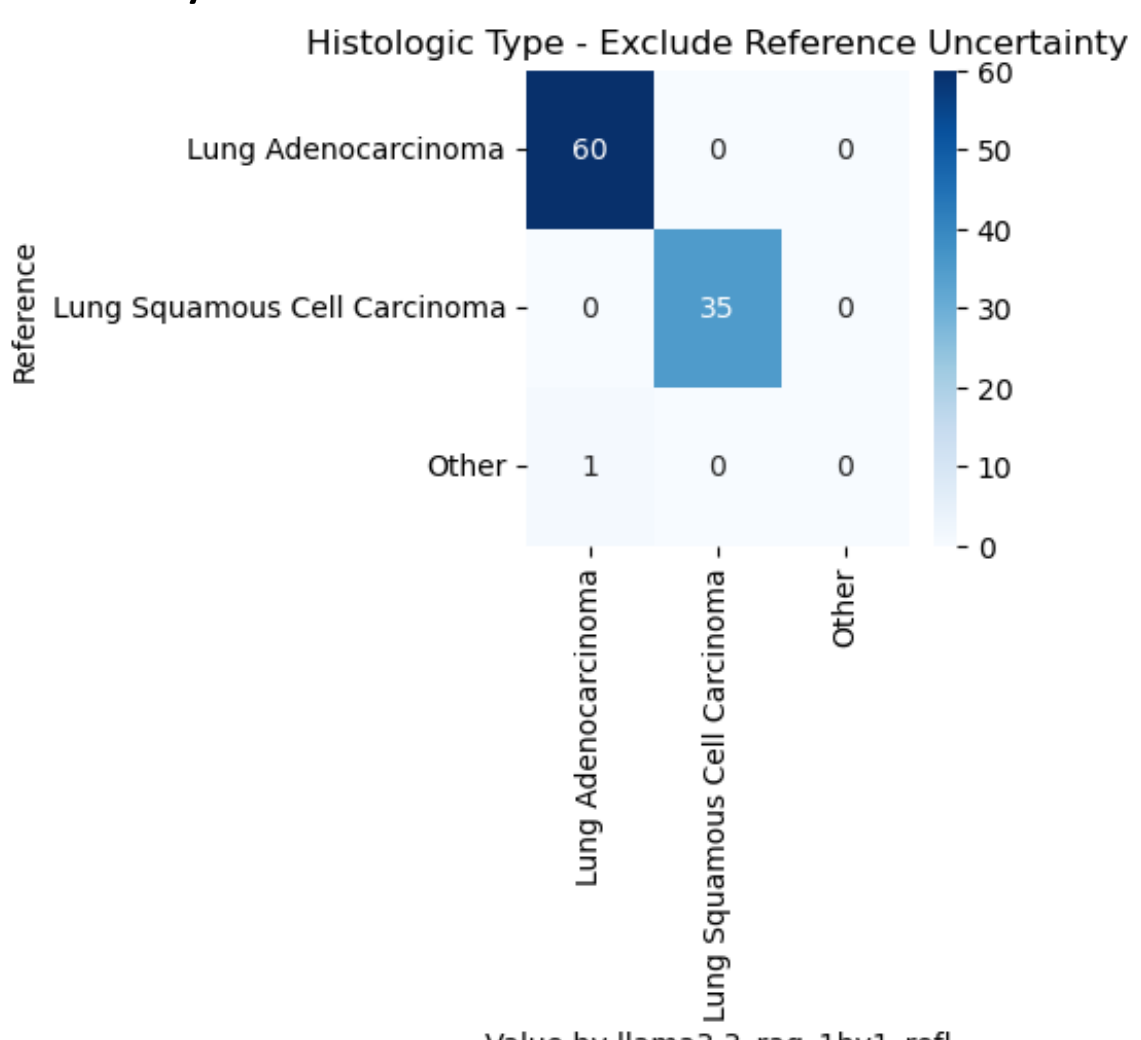

Accuracy: 0.9896

Accuracy: 0.9896

Comparing confusion matrixes of key variables in initial TNM estimation and the result after 15 rounds of deep reflective reasoning with backtracking 10 steps. At round 15, most cases converged in deep reflective reasoning and only 2 cases left unconverged. The performance of Tumor Size and Histologic Type remain unchanged, holding correct rate of 0.964 and accuracy of 0.9896 respectively. The core variables have improved accuracy, pT from 0.8421 to 0.8842, pN from 0.8854 to 0.9479, and Tumor Stage from 0.6795 to 0.8333.

## Fig. S4 Deep reflective reasoning for CD99 pattern

a. Initial CD99 pattern identification

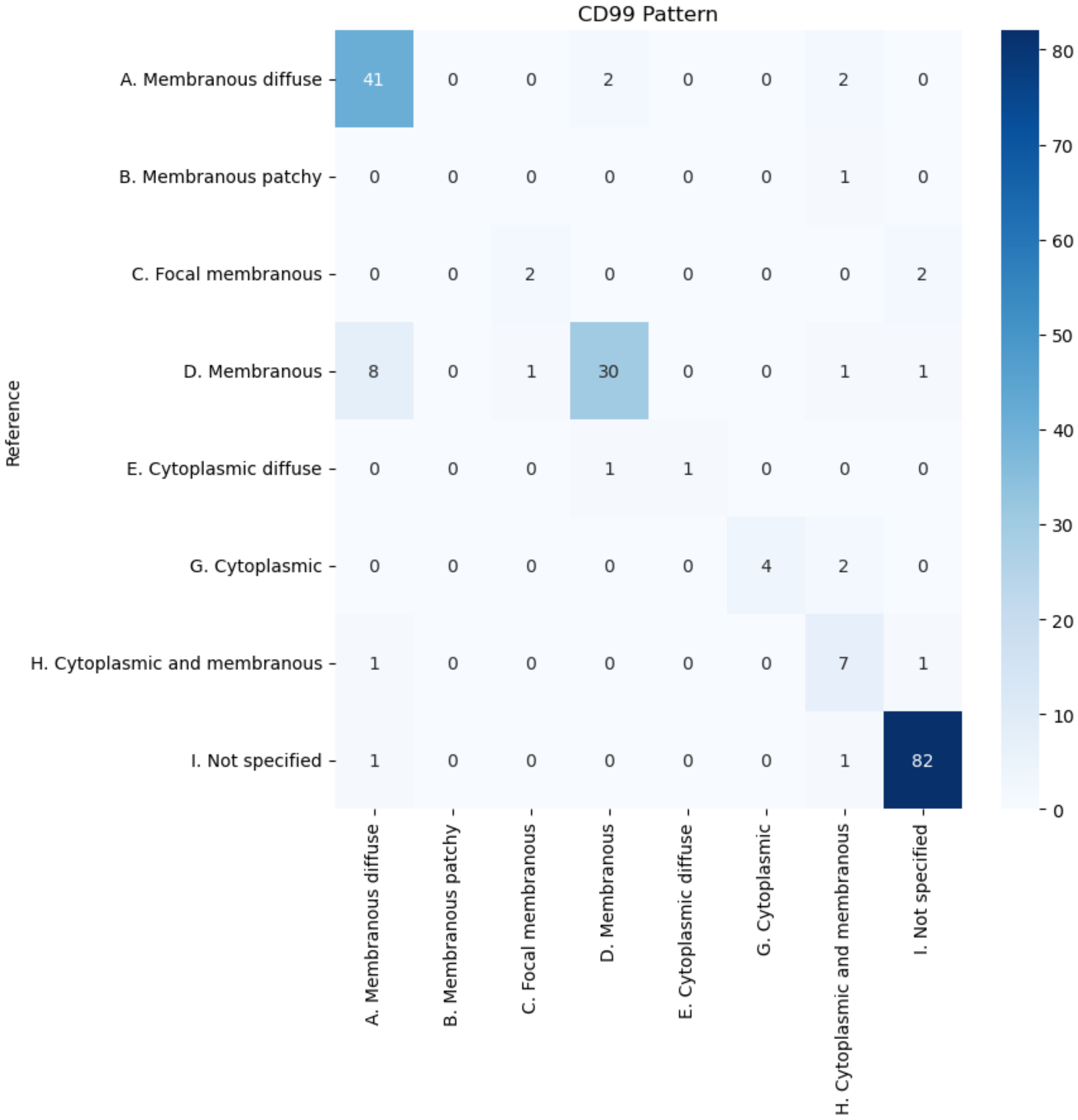

Accuracy: 0.8698
95% CI: [0.8177, 0.9115]
F1 (weighted): 0.8678
Recall (weighted): 0.8698
Precision (weighted): 0.8787

b. Converged at round 3 with backtracking 2

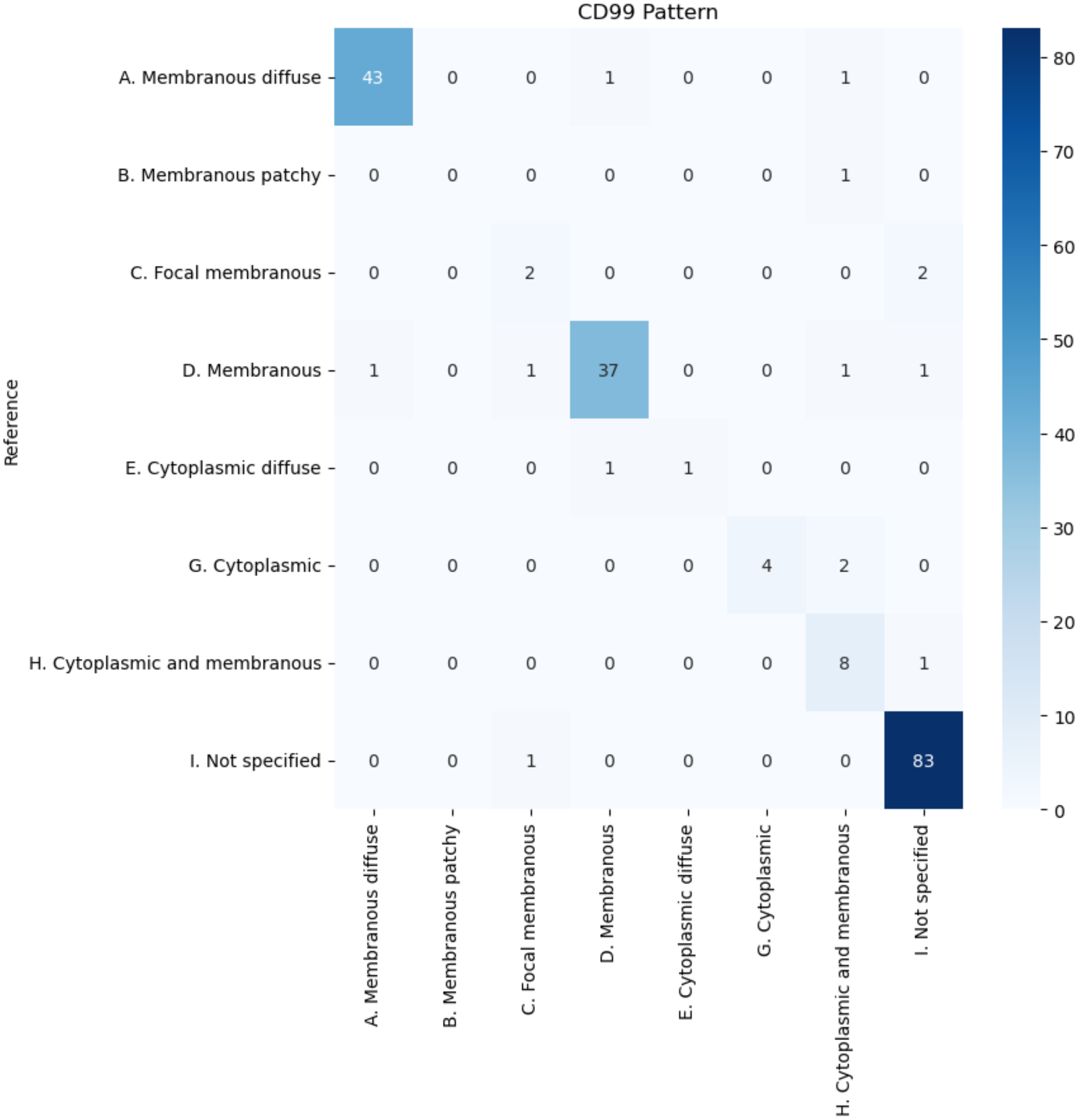

Accuracy: 0.9271
95% CI: [0.8906, 0.9635]
F1 (weighted): 0.9252
Recall (weighted): 0.9271
Precision (weighted): 0.93

a. After round 15 with backtracking 3

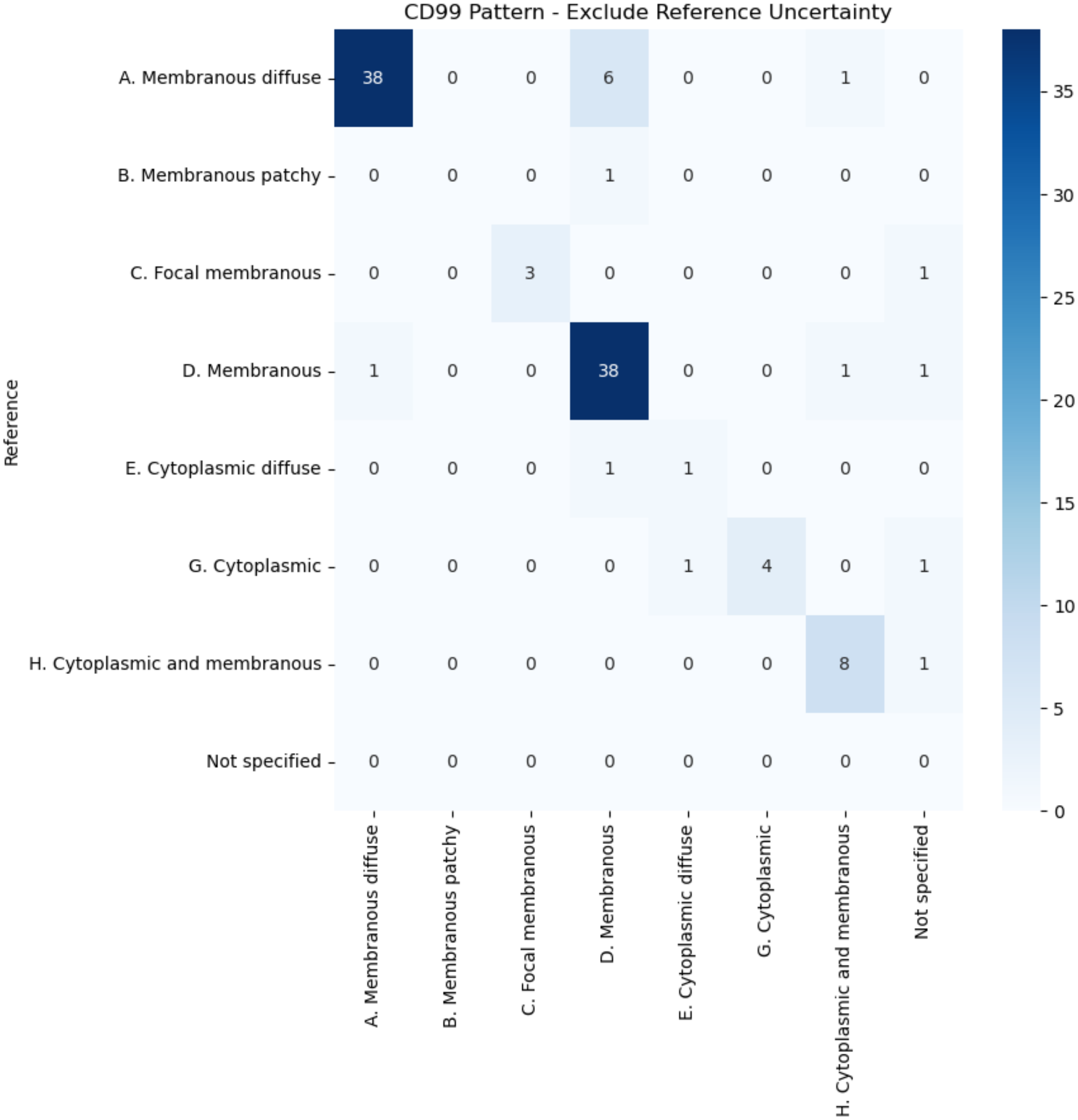

Accuracy: 0.8519
95% CI: [0.7778, 0.9167]
F1 (weighted): 0.8642
Recall (weighted): 0.8519
Precision (weighted): 0.8881

# Fig. S5 Lung cancer TNM staging AJCC7 rules

## DEFINITIONS OF TNM

***Primary Tumor (T)***

| | |
|---|---|
| TX | Primary tumor cannot be assessed, or tumor proven by the presence of malignant cells in sputum or bronchial washings but not visualized by imaging or bronchoscopy |
| T0 | No evidence of primary tumor |
| Tis | Carcinoma in situ |
| T1 | Tumor 3 cm or less in greatest dimension, surrounded by lung or visceral pleura, without bronchoscopic evidence of invasion more proximal than the lobar bronchus (i.e., not in the main bronchus)* |
| T1a | Tumor 2 cm or less in greatest dimension |
| T1b | Tumor more than 2 cm but 3 cm or less in greatest dimension |
| T2 | Tumor more than 3 cm but 7 cm or less or tumor with any of the following features (T2 tumors with these features are classified T2a if 5 cm or less); Involves main bronchus, 2 cm or more distal to the carina; Invades visceral pleura (PL1 or PL2); Associated with atelectasis or obstructive pneumonitis that extends to the hilar region but does not involve the entire lung |
| T2a | Tumor more than 3 cm but 5 cm or less in greatest dimension |
| T2b | Tumor more than 5 cm but 7 cm or less in greatest dimension |
| T3 | Tumor more than 7 cm or one that directly invades any of the following: parietal pleural (PL3), chest wall (including superior sulcus tumors), diaphragm, phrenic nerve, mediastinal pleura, parietal pericardium; or tumor in the main bronchus (less than 2 cm distal to the carina* but without involvement of the carina; or associated atelectasis or obstructive pneumonitis of the entire lung or separate tumor nodule(s) in the same lobe |
| T4 | Tumor of any size that invades any of the following: mediastinum, heart, great vessels, trachea, recurrent laryngeal nerve, esophagus, vertebral body, carina, separate tumor nodule(s) in a different ipsilateral lobe |

*The uncommon superficial spreading tumor of any size with its invasive component limited to the bronchial wall, which may extend proximally to the main bronchus, is also classified as T1a.

***Regional Lymph Nodes (N)***

| | |
|---|---|
| NX | Regional lymph nodes cannot be assessed |
| N0 | No regional lymph node metastases |
| N1 | Metastasis in ipsilateral peribronchial and/or ipsilateral hilar lymph nodes and intrapulmonary nodes, including involvement by direct extension |
| N2 | Metastasis in ipsilateral mediastinal and/or subcarinal lymph node(s) |
| N3 | Metastasis in contralateral mediastinal, contralateral hilar, ipsilateral or contralateral scalene, or supraclavicular lymph node(s) |

***Distant Metastasis (M)***

| | |
|---|---|
| M0 | No distant metastasis |
| M1 | Distant metastasis |
| M1a | Separate tumor nodule(s) in a contralateral lobe tumor with pleural nodules or malignant pleural (or pericardial) effusion* |
| M1b | Distant metastasis (in extrathoracic organs) |

From Goldstraw P, Crowley J, Chansky K, et al.: The IASLC Lung Cancer Staging Project: Proposals for the revision of the TNM stage groupings in the forthcoming (seventh) edition of the TNM classification of malignant tumours. *J Thorac Oncol* 2:706–714, 2007, with permission.

*Most pleural (and pericardial) effusions with lung cancer are due to tumor. In a few patients, however, multiple cytopathologic examinations of pleural (pericardial) fluid are negative for tumor, and the fluid is nonbloody and is not an exudate. Where these elements and clinical judgment dictate that the effusion is not related to the tumor, the effusion should be excluded as a staging element and the patient should be classified as M0.

## ANATOMIC STAGE/PROGNOSTIC GROUPS

| | | | |
|---|---|---|---|
| Occult carcinoma | TX | N0 | M0 |
| Stage 0 | Tis | N0 | M0 |
| Stage IA | T1a | N0 | M0 |
| | T1b | N0 | M0 |
| Stage IB | T2a | N0 | M0 |
| Stage IIA | T2b | N0 | M0 |
| | T1a | N1 | M0 |
| | T1b | N1 | M0 |
| | T2a | N1 | M0 |
| Stage IIB | T2b | N1 | M0 |
| | T3 | N0 | M0 |
| Stage IIIA | T1a | N2 | M0 |
| | T1b | N2 | M0 |
| | T2a | N2 | M0 |
| | T2b | N2 | M0 |
| | T3 | N1 | M0 |
| | T3 | N2 | M0 |
| | T4 | N0 | M0 |
| | T4 | N1 | M0 |
| Stage IIIB | T1a | N3 | M0 |
| | T1b | N3 | M0 |
| | T2a | N3 | M0 |
| | T2b | N3 | M0 |
| | T3 | N3 | M0 |
| | T4 | N2 | M0 |
| | T4 | N3 | M0 |
| Stage IV | Any T | Any N | M1a |
| | Any T | Any N | M1b |

## Fig. S6 Deep reflective reasoning algorithm

```
Algorithm: Deep reflective reasoning
Inputs:
prompt_template
m – maximum rounds of reflections.
p – number of back-views (last p rounds of reflections to review)
case_set – the collection of all cases, including id and input data.
base_pred – the base prediction (variable-value pairs) of all cases.
Output:
result – the final predictions for all cases, as result of the reflections.
refl_seq – the sequence of reflections

# initialize
refl_seq = base_pred
for case i in case_set:
    case_converged[i]=0
    # 0 denotes not converged; k denotes reflection converged at round k

# reflection
for k = 1 to m:
    # reflection round k:
    cases_updated = 0
    for case i in case_set:
        if case_converged[i] == 0:
            # from refl_seq, get case i's last p rounds of reflections, in ordere of k-1, ..., k-p
            v_p = get_record(refl_seq,i,p)
            prompt <- construct prompt from prompt_template, case i data, and v_p
            response = query_llm(prompt)
            (v,c) = transform_to_record(response) # v: new value vector; c: critique and explanation
            if v==v_p[0]: # if new value vector is eqivalent to the value vector in the last round
                case_converged[i] = k
            else:
                refl_seq <- appeend (v,c) to refl_seq
                cases_updated += 1
    if cases_updated == 0:
        result <- get the latest output for all cases from refl_seq
        break
return result, refl_seq
```